\documentclass[]{bytedance_seed}

\usepackage[toc,page,header]{appendix}

\usepackage{minitoc}
\usepackage{enumitem}
\newcommand{\ourbenchmark}{MARS-Bench}
\usepackage{pifont}
\usepackage{amssymb}
\usepackage{makecell}

\newcommand{\redmark}{\textcolor{red}{\ding{55}}}

\newcommand{\greencheck}{\textcolor{green}{\ding{51}}}
\usepackage{xcolor}
\usepackage{booktabs}
\usepackage{hyperref}

\usepackage{booktabs}
\usepackage{hyperref}
\definecolor{first}{RGB}{200,240,255}    
\newcommand{\first}[1]{\cellcolor{first}{#1}}
\newcommand{\second}[1]{\textbf{#1}}
\newcommand{\third}[1]{\underline{#1}}
\usepackage{graphicx}
\usepackage{subcaption}
\usepackage{array}

\usepackage{latexsym}

\usepackage{tabularx}
\usepackage{graphicx} 
\usepackage{booktabs}
\usepackage{multirow} 
\usepackage{colortbl} 
\usepackage{tabularx}
\usepackage{makecell}
\usepackage{booktabs}
\usepackage{listings}

\usepackage{tcolorbox}
\tcbuselibrary{breakable, skins}
\usepackage{tikz}

\newtcolorbox[auto counter, number within=section]{Prompt}[2][]{%
  colback=white, 
  colframe=cyan, 
  width=\textwidth, 
  arc=5mm, 
  boxrule=0.8mm, 
  title=\large #2, 
  breakable, 
  fonttitle=\small, 
  fontupper=\footnotesize, 
  #1 
}

\newtcolorbox[auto counter, number within=section]{QuestionCase}[2][]{%
  colback=white,
  colframe=yellow!50!red,
  width=\textwidth, 
  arc=5mm, 
  boxrule=0.8mm, 
  title=\large #2, 
  breakable, 
  fonttitle=\small, 
  fontupper=\footnotesize, 
  #1 
}
\usepackage{graphicx}
\usepackage{float}
\usepackage{wrapfig}

\title{MARS-Bench: A Multi-turn Athletic Real-world Scenario Benchmark for Dialogue Evaluation}

\author[1,2,*,\ddagger]{Chenghao Yang}
\author[2,*]{Yinbo Luo}
\author[1,\dagger]{Zhoufutu Wen}
\author[2,\dagger]{Qi Chu}
\author[2]{Tao Gong}
\author[1]{Longxiang Liu}
\author[1]{Kaiyuan Zhang}
\author[1]{Jianpeng Jiao}
\author[1]{Ge Zhang}
\author[1]{Wenhao Huang}
\author[2]{Nenghai Yu}
\affiliation[1]{ByteDance Seed}
\affiliation[2]{University of Science and Technology of China}

\contribution[*]{Equal contribution}
\contribution[\dagger]{Corresponding authors}
\contribution[\ddagger]{Work done at ByteDance Seed}

\abstract{
Large Language Models (\textbf{LLMs}), e.g. ChatGPT, have been widely adopted in real-world dialogue applications. 
However, LLMs' robustness, especially in handling long complex dialogue sessions, including frequent motivation transfer, sophisticated cross-turn dependency, is criticized all along. 
Nevertheless, no existing benchmarks can fully reflect these weaknesses.
We present \textbf{MARS-Bench}, a \textbf{M}ulti-turn \textbf{A}thletic \textbf{R}eal-world \textbf{S}cenario Dialogue \textbf{Bench}mark, designed to remedy the gap.
MARS-Bench is constructed from play-by-play text commentary so to feature realistic dialogues specifically designed to evaluate three critical aspects of multi-turn conversations: Ultra Multi-turn, Interactive Multi-turn, and Cross-turn Tasks.
Extensive experiments on MARS-Bench also reveal that closed-source LLMs significantly outperform open-source alternatives, explicit reasoning significantly boosts LLMs' robustness on handling long complex dialogue sessions, and LLMs indeed face significant challenges when handling motivation transfer and sophisticated cross-turn dependency.
Moreover, we provide mechanistic interpretability on how attention sinks due to special tokens lead to LLMs' performance degradation when handling long complex dialogue sessions based on attention visualization experiment in Qwen2.5-7B-Instruction.
}

\date{\today}
\correspondence{Zhoufutu Wen at \email{liniuniu@bytedance.com}, Qi Chu at \email{qchu@ustc.edu.cn}}

\checkdata[Project Page]{\url{https://syuchin.github.io/MARS-Bench/}}

\begin{document}
\maketitle


\section{Introduction}
\label{sec: intro}

\begin{figure}[t]
    \centering
    \includegraphics[width=\textwidth]{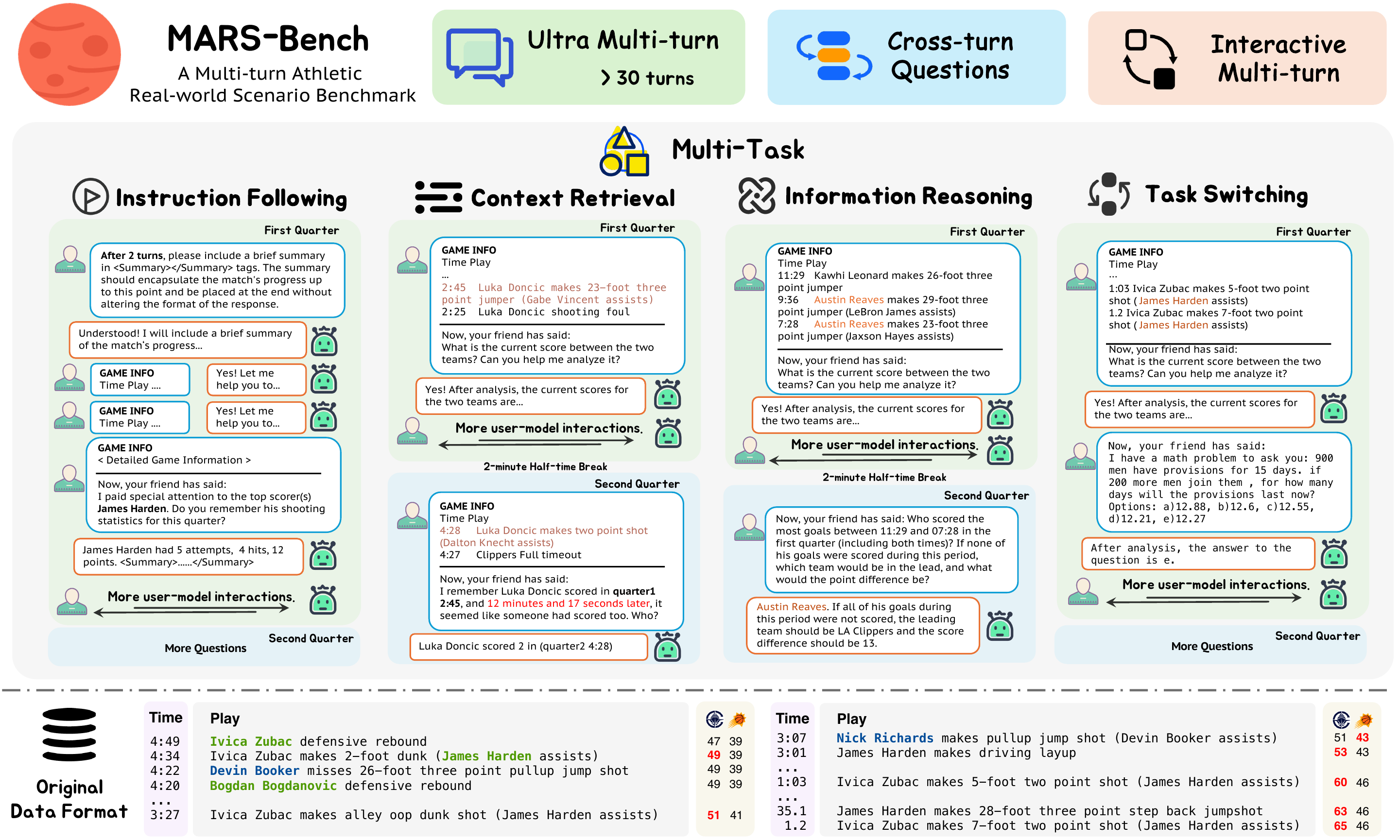}
    \caption{\textbf{Overview of \ourbenchmark{}.} \ourbenchmark{} is constructed from real-world play-by-play sports data and supports Ultra Multi-turn (UMT), Cross-turn Tasks (CTT), and Interactive Multi-turn (IMT) dialogue scenarios. It comprises four core task categories—Instruction Following, Context Retrieval, Information Reasoning, and Task Switching—each illustrated with representative dialogue examples. The bottom section shows the structured game data format.}
    \label{fig:benchmark_overview}
\end{figure}

Large Language Models (\textbf{LLMs}) have made remarkable advances, enabling fluent interactions with users, even on sessions with more than 30 turns, messy information, and unnatural motivation transfer.
However, LLMs' robustness has long been criticized when handling users' shifts between information‑seeking questions, reasoning tasks, and creative content generation in same dialogue sessions without clear task boundaries, while no existing evaluations fully reflect the weakness.
We point out that LLMs must (i) retrieve evidence dispersed across distant, sometimes dozens-of-turn-old, utterances, and (ii) reason jointly over these fragments while adapting to frequent task switches to handle the aforementioned scenarios.
As these conversational scenarios grow increasingly complex, there is a clear need for robust evaluation protocols that can systematically assess LLMs’ ability to understand, reason, and respond coherently across conversational turns.

\begin{figure*}[t]
    \centering
    \begin{subfigure}[t]{0.48\textwidth}
        \centering
        \includegraphics[width=\linewidth]{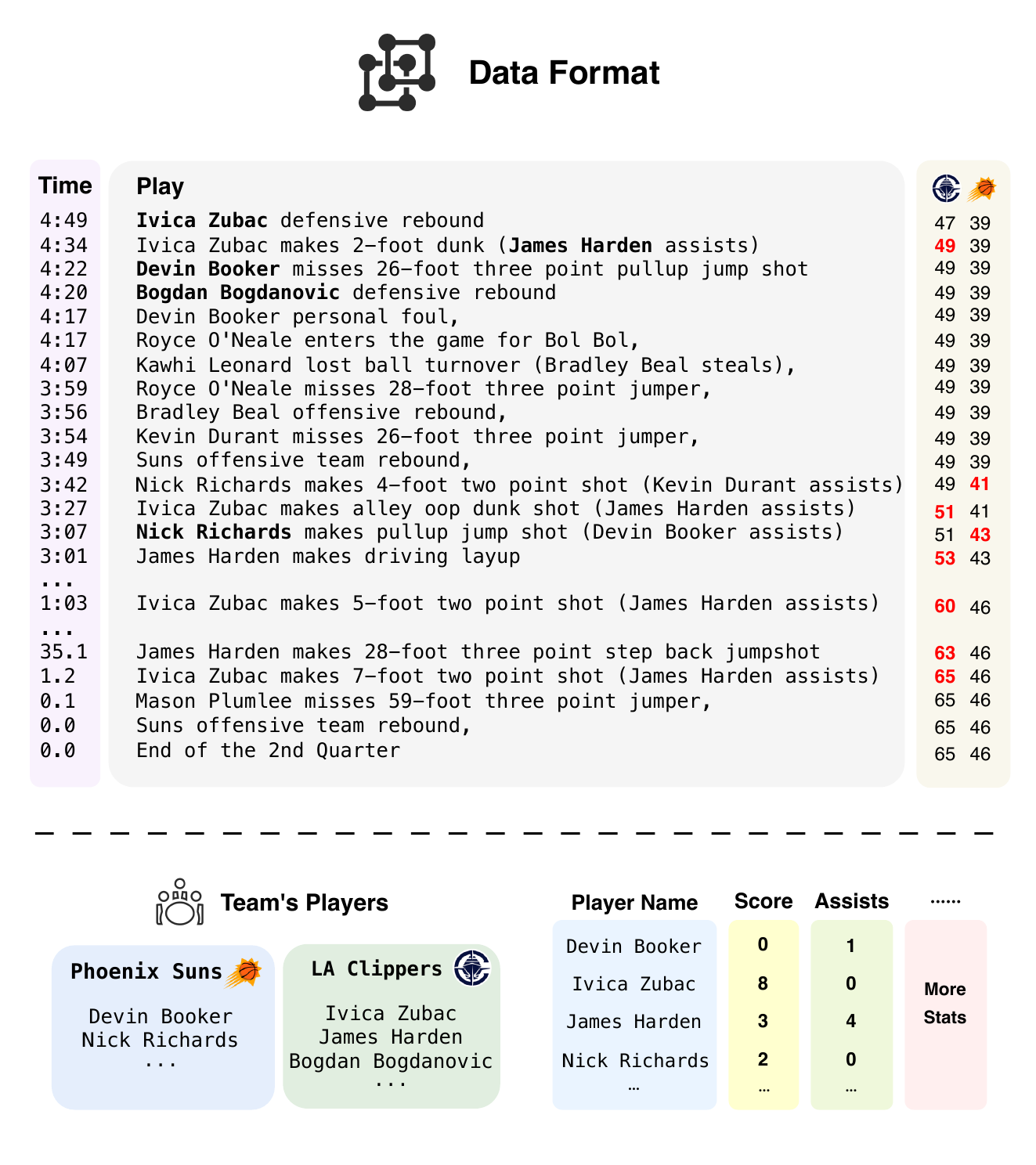} 
        \caption{\textbf{Overview of the Data Format.} Each sample includes (1) play-by-play records, (2) team rosters, and (3) player statistics. The first two are model inputs; the third is used for answer verification.}
        \label{fig:data_format}
    \end{subfigure}
    \hfill
    \begin{subfigure}[t]{0.48\textwidth}
        \centering
        \includegraphics[width=\linewidth]{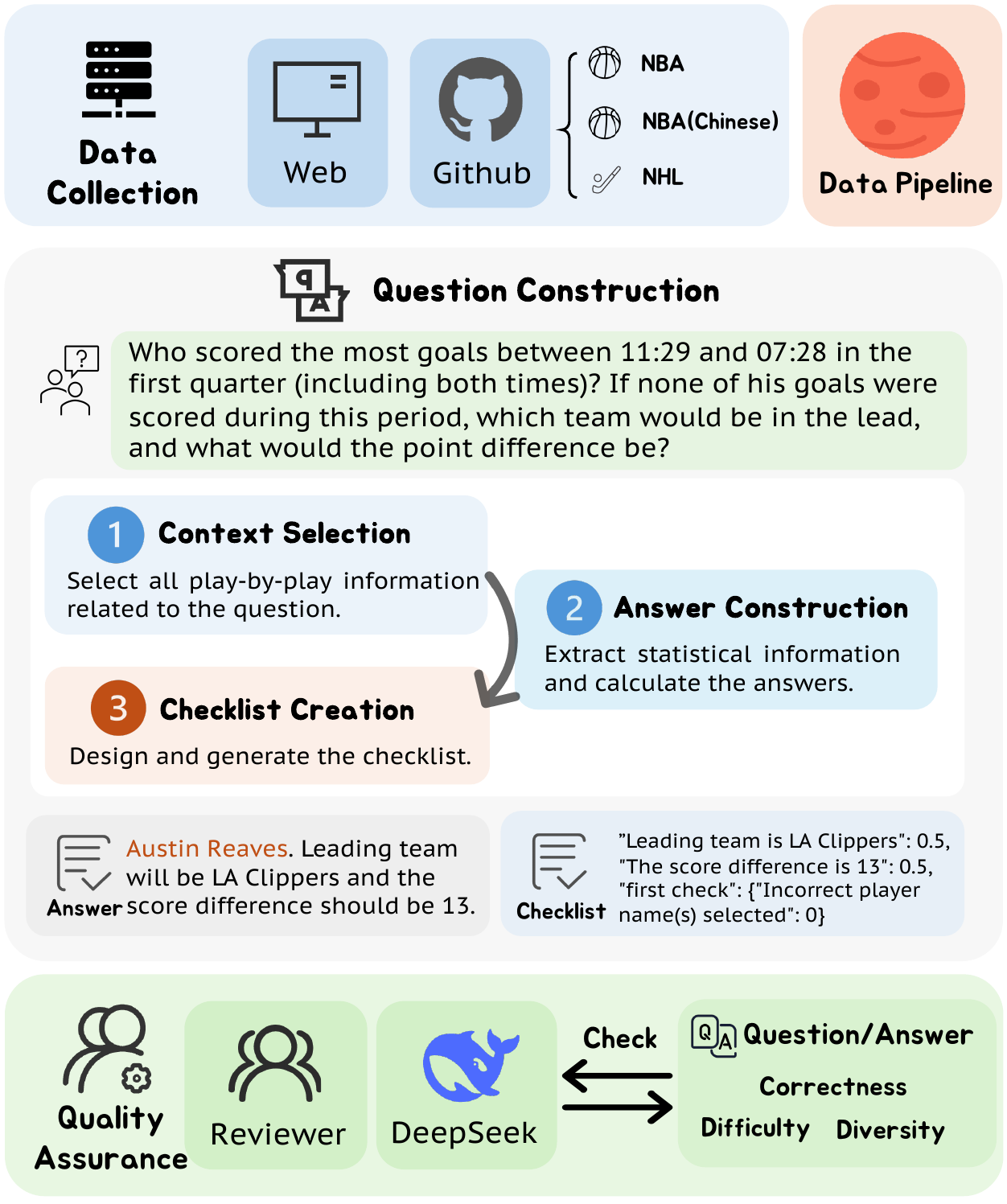} 
        \caption{\textbf{Overview of the Data Construction Pipeline.} The pipeline comprises three stages: (1) \textbf{Data Collection}, where sports data is gathered; (2) \textbf{Question Construction}, including (Question, Answer, Checklist) generation; and (3) \textbf{Quality Assurance} via review for correctness and difficulty.}
        \label{fig:data_pipeline}
    \end{subfigure}
    \caption{\textbf{Overview of the dataset construction and format.} Subfigure~(a) shows a sample structure; subfigure~(b) shows the pipeline.}
    \label{fig:data_overview}
\end{figure*}

However, many benchmarks focus on short conversations, provide the full dialogue history upfront rather than revealing it turn by turn, and rarely test reasoning over information scattered across distant turns~\citep{MT-Bench-101, MT-Bench, MT-Eval, Parrot, MultiChallenge}. 
Agent-based benchmarks~\citep{AgentBench, RealWebAssist} explore complex task settings, but often lack alignment with real-world dialogue scenarios, making it difficult to assess how models adapt in nature.
These limitations highlight the lack of comprehensive benchmarks for Interactive Multi-turn (IMT) dialogue, Cross-turn Tasks (CTT), and Ultra Multi-turn (UMT) scenarios.

To address gaps in existing research, we propose \textbf{MARS-Bench}, a multi-turn dialogue benchmark constructed from real-world play-by-play sports data. 
MARS-Bench emphasizes three key features: \textbf{Ultra Multi-turn Dialogues} with over 30 turns per instance, capturing instruction shifts and contextual evolution; \textbf{Cross-turn Tasks} that require reasoning over non-adjacent information; and \textbf{Interactive Multi-turn Generation}, where LLMs must respond at every turn, reflecting realistic user interactions and frequent task switches.
Built on top of these settings, MARS-Bench defines four core tasks: instruction following, context retrieval, information reasoning, and task switching. 
These tasks jointly enable a comprehensive evaluation of multi-turn and multi-task dialogue capabilities.
Extensive experiments on state-of-the-art LLMs with MARS-Bench reveal that:

\begin{itemize}[leftmargin=*]
\item \textbf{Closed-Source LLMs Leaders}: Closed-source models, e.g. Claude-3.7-Sonnet-Thinking,  substantially outperform open-source alternatives, particularly in tasks requiring deep contextual understanding and multi-turn reasoning.

\item \textbf{LLMs Benefit from Explicit Reasoning}: Models employing explicit reasoning mechanisms (\textit{System 2}) consistently achieve higher accuracy and stability, whereas heuristic-driven (\textit{System 1}) approaches falter with increased task complexity.

\item \textbf{LLMs Struggle with Multi-Turn Dialogue}: Models face notable difficulties in instruction following, retaining context across multiple turns, and managing cumulative errors from incremental predictions, highlighting critical bottlenecks in long-range contextual memory and structured inference processes.
\end{itemize}

\section{\texttt{\ourbenchmark{}}: Design and Construction}
\label{sec: method}
This section is organized as follows: Data Collection and Processing~(\ref{subsec: data_construction_process}), Task Categories~(\ref{subsec: dataset_categories}), and Benchmark Statistics~(\ref{subsec: statistic}).

\subsection{Data Collection and Processing}
\label{subsec: data_construction_process}

Building on the play-by-play textual game data, we construct a three-stage data pipeline comprising data collection, question construction, and manual verification, as illustrated in Figure~\ref{fig:data_pipeline}.

\paragraph{\textbf{Data Collection}}

Sports play-by-play records provide event sequences with temporal order and reliable statistical information, which enable objective and consistent evaluation. Building on this advantage, we collect English play-by-play records and post-game statistics for the NBA and NHL from ESPN, and supplement them with Chinese NBA data from Hupu, a major sports community in China, to enhance both linguistic and domain diversity. The collected data is standardized into structured formats for multi-turn dialogue modeling, with each sample corresponding to a complete sports game. Figure~\ref{fig:data_format} illustrates the structure of the collected data.

\paragraph{\textbf{Question Construction}}
Each question is manually designed according to a specific task type. The construction process begins with official play-by-play records collected from ESPN and Hupu. Based on these records and the pre-defined question types summarized in Table~\ref{tab:task_categories}, a set of distinct questions is systematically constructed for each game. For each instance, relevant information is extracted from the play-by-play entries using regular expressions, from which the unique answer can be directly derived. A corresponding checklist is then created, which specifies objective statistical evidence (e.g., scores and timestamps) as the verification criteria, thereby ensuring answer correctness and enabling its use in subsequent automatic validation. Together, these steps yield a set of \textit{(question, answer, checklist)} triplets that form the basis for evaluating model responses.

\paragraph{\textbf{Quality Assurance}}  
All annotation and verification work is conducted by the authors. To ensure the correctness, difficulty, and diversity of the constructed \textit{(Question, Answer, Checklist)} triplets, we implement a three-stage quality control process:  
(1) \textit{Sampling verification}, where five representative questions per type are manually checked against standardized play-by-play records to confirm the reliability of automatic answer extraction;  
(2) \textit{Cross verification}, where both authors independently review the generated outputs to reduce subjective bias and improve overall coverage;  
(3) \textit{Consistency verification}, where a large language model (DeepSeek-V3-0324) is employed to assist human reviewers in identifying potential omissions and underspecified outputs for further inspection.

\begin{table}[h]
\renewcommand{\arraystretch}{1.4}
\centering
\caption{\textbf{Task categories and subtask types in \ourbenchmark{}}, along with the number of subtasks generated per game segment. Each segment corresponds to a natural period in sports games—three periods in the NHL and four quarters in the NBA (excluding overtime).}
\label{tab:task_categories}
\vskip 0.1in
\setlength{\tabcolsep}{6pt}
\resizebox{\textwidth}{!}{
\begin{tabular}{@{}m{3.8cm}m{5cm}m{4.5cm}m{6.8cm}c@{}}
\toprule
\textbf{Task Type} & \textbf{Description} & \textbf{Sub-task} & \textbf{Sub-task Description} & \textbf{Questions per Period} \\
\midrule
\multirow{3}{*}{Instruction Following} & \multirow{3}{\dimexpr5cm-2\tabcolsep}{Follow turn-specific instructions with format constraints.}
& Fixed-format Single-turn Response & Follow the format specified for the current dialogue turn. & 1 \\
\cmidrule(lr){3-5}
& & Turn-conditioned Prompted Formatting & Adapt the response format according to system instructions at each turn. & 8 \\
\cmidrule(lr){3-5}
& & Turn-conditioned Inferred Formatting & Adjust the response format based on instructions inferred from prior dialogue turns. & 1 \\

\cmidrule(lr){1-5}
\multirow{2}{*}{Context Retrieval} & \multirow{2}{\dimexpr5cm-2\tabcolsep}{Locate and retrieve factual information from previous dialogue turns.}
& Anchored Event Retrieval & Given a time anchor and interval, retrieve a specific event. & 2 \\
\cmidrule(lr){3-5}
& & Interval-based Event Retrieval & Given a start and end time, retrieve events of a specific type. & 1 \\
\cmidrule(lr){1-5}

\multirow{3}{*}{Information Reasoning} & \multirow{3}{\dimexpr5cm-2\tabcolsep}{Aggregate and reason over distributed contextual information.}
& Current Score Tracking & Provide the current score for both teams. & 1 (last period) \\
\cmidrule(lr){3-5}
& & Score Lead Fluctuation Detection & Identify the number and timing of score lead changes between the two teams within a given time period. & 1 \\
\cmidrule(lr){3-5}
& & Player Performance Impact Analysis & Given a time span, analyze how a change in a player's performance affected the game situation. & 2 \\
\cmidrule(lr){1-5}

\multirow{2}{*}{Task Switching} & \multirow{2}{\dimexpr5cm-2\tabcolsep}{Handle abrupt interleaving of unrelated queries.}
& In-context Reasoning Query & Ask questions related to the match. & 3 \\
\cmidrule(lr){3-5}
& & Out-of-context Math Query & Ask unrelated mathematical questions from MathQA~\citep{MathQA}. & 3 \\
\bottomrule
\end{tabular}
}
\end{table}
\subsection{Task Categories}
\label{subsec: dataset_categories}

Grounded in systematic observations from real-world dialogue applications, we identify three core capabilities required for robust multi-turn dialogue modeling: consistent instruction following across turns, effective retrieval and reasoning over long-range context, and robustness to unrelated or interleaved task inputs. To operationalize these requirements under ultra multi-turn settings, we categorize model behaviors into four complementary dimensions: \textbf{Instruction Following}, which measures the ability to track turn-level structures and accurately execute user instructions; \textbf{Context Retrieval}, which assesses the extraction of salient information from extended dialogue histories; \textbf{Information Reasoning}, which captures the model’s capacity to integrate retrieved context for coherent, semantically grounded reasoning; and \textbf{Task Switching}, which evaluates adaptability to abrupt task shifts and stability under heterogeneous objectives. These dimensions collectively constitute a structured framework for evaluating interactive multi-turn (IMT) dialogue, cross-turn tasks (CTT), and ultra multi-turn (UMT) scenarios.

Each game is divided into periods (e.g., NBA quarters, NHL periods), with each period split into five score-tracking turns that form a multi-turn dialogue. Category-specific questions are inserted at appropriate points based on the four task categories. Models must respond incrementally and maintain coherence throughout.

Figure~\ref{fig:benchmark_overview} provides abstract illustrations of the task categories, while Table~\ref{tab:task_categories} outlines their descriptions, subtask distributions, and instance counts per period. Full task examples and question placements are detailed in Appendix~\ref{appendix: task categories}.

\begin{table}[h]
\renewcommand{\arraystretch}{1.2}
\setlength{\extrarowheight}{1pt}
\centering
\caption{Comparison of \ourbenchmark{} with Other Multi-turn Dialogue Benchmarks}
\label{tab:benchmark_comparison}
\vskip 0.05in
\resizebox{0.95\columnwidth}{!}{
\begin{tabular}{lcccccc}
\toprule
\textbf{Benchmark} & \textbf{Real Interaction} & \textbf{Cross-turn} & \textbf{Multi-task} & \textbf{Avg. Turns} & \textbf{Total Queries} & \textbf{Language} \\
\midrule
MT-Bench         & \redmark   & \redmark   & \redmark   & 1--2  & 80    & English \\
MT-Bench-101     & \redmark   & \greencheck & \greencheck & 3     & 4208  & English \\
MultiChallenge   & \redmark   & \greencheck & \greencheck & 5     & 1365  & English \\
\ourbenchmark{} (\textbf{Ours}) & \greencheck & \greencheck & \greencheck & 33    & 4010  & English, Chinese \\
\bottomrule
\end{tabular}
}
\end{table}






\noindent
\begin{minipage}{0.45\textwidth}
\subsection{Benchmark Statistics}
\label{subsec: statistic}

\ourbenchmark{} consists of 120 games, with an average of 33.42 dialogue turns per game.  
Each of the four task categories comprises 30 games, evenly distributed across three sports domains: 15 NBA (English), 10 NBA (Chinese), and 5 NHL (English).  

Figure~\ref{fig:stats} visualizes the distribution of tasks and domains, where the outer ring indicates task categories and the inner ring reflects domain composition.

Table~\ref{tab:benchmark_comparison} presents a comparison between \ourbenchmark{} and existing multi-turn dialogue benchmarks. In contrast to previous datasets, \ourbenchmark{} offers substantially longer dialogues, incorporates real user–model interactions, and covers a more diverse set of tasks spanning multiple languages and domains.
\end{minipage}%
\hfill 
\begin{minipage}{0.5\textwidth}
\centering
\includegraphics[width=\linewidth]{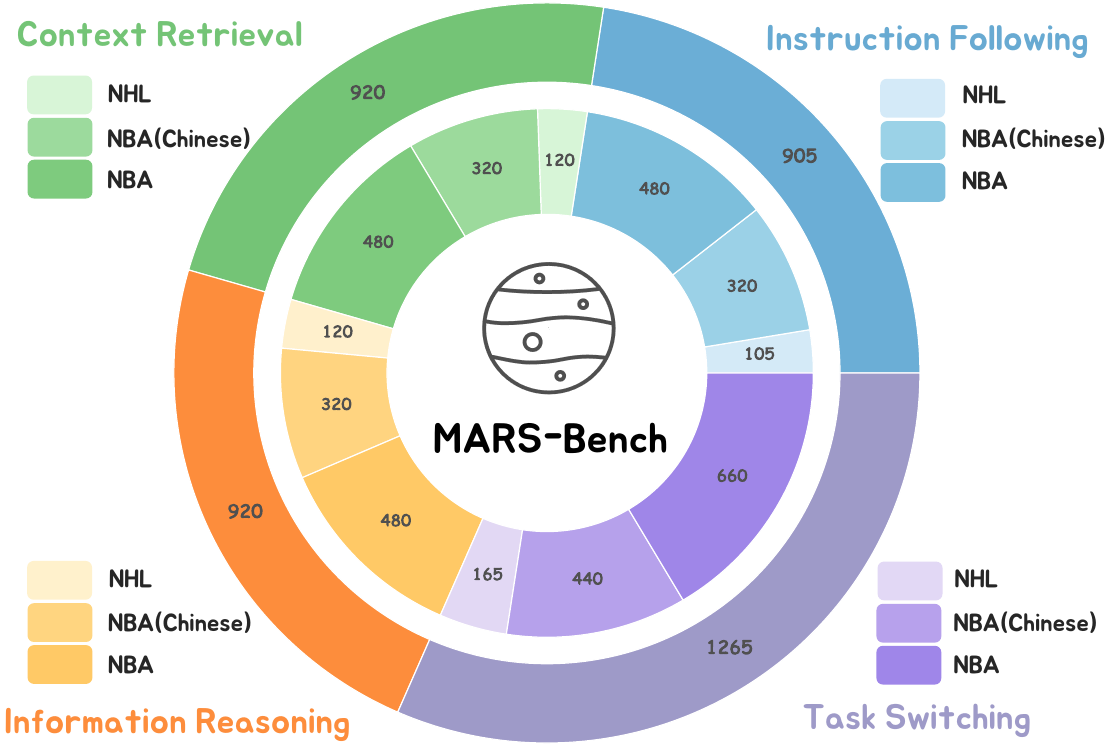}
\captionof{figure}{\textbf{Statistics of \ourbenchmark{}.} The outer ring shows the distribution of the four task categories, and the inner ring indicates the corresponding sports domains: NBA (English), NBA (Chinese), and NHL (English). Numbers on the chart represent the total number of model interaction turns per task.}
\label{fig:stats}
\end{minipage}

\begin{table}[h!]
\renewcommand{\arraystretch}{1.2}
\setlength{\extrarowheight}{3pt}
\centering
\caption{\textbf{Performance of different models on \ourbenchmark{}.} The benchmark includes four task categories: Instruction Following (IF), Context Retrieval (CR), Information Reasoning (IR), and Task Switching (TS). \first{Shaded cells} indicate the best performance, \second{bold} indicates the second-best, and \third{underlined} the third-best. For more details about the scores of math questions, please refer to the Appendix~\ref{appendix: results with other type}.}
\label{tab:main_result}
\vskip 0.15in
\small
\resizebox{0.92\textwidth}{!}{
\begin{tabular}{lccccccc}
\toprule
{\textbf{Model}} & {\textbf{Reasoning}} & {\textbf{Open Source}} &{\textbf{Overall}} & {\textbf{IF}} & {\textbf{CR}} & {\textbf{IR}} & {\textbf{TS}} \\
\midrule
Gemini-2.5-Pro~\citep{google_gemini25pro0506} & \greencheck & \redmark & \first{72.44} & \first{65.08} & \first{87.06} &\first{70.92} & \second{66.72} \\
Claude-3.7-Sonnet-Thinking~\citep{anthropic_claude37sonnet} & \greencheck & \redmark & \second{62.29} & 43.28 & 71.51 & \second{66.98} & \first{67.38} \\
o1-1217~\citep{openai_o11217} & \greencheck & \redmark & 59.62 & 53.09 & 64.48 & \third{62.63} & 58.28 \\
Gemini-2.5-Flash~\citep{google_gemini25flash} & \greencheck & \redmark & 59.22 & 45.96 & \second{77.76} & 52.93 & \third{60.23} \\
GPT-4.5-Preview~\citep{openai_gpt45preview} & \redmark & \redmark & 53.33 & \third{55.52} & 66.65 & 50.43 & 40.74 \\
Doubao-1.5-Pro-Thinking~\citep{bytedance_doubao15prothinking} & \greencheck & \redmark & 52.62 & 51.99 & 55.64 & 52.17 & 50.69 \\
Grok3~\citep{grok3} & \greencheck & \redmark & 51.21 & \second{61.19} & \third{73.91} & 33.89 & 35.87 \\
o4-mini-0416~\citep{openai_o4mini} & \greencheck & \redmark & 47.13 & 47.48 & 61.26 & 39.74 & 40.03 \\
DeepSeek-R1~\citep{deepseek_r1} & \greencheck & \greencheck & 45.42 & 53.04 & 49.23 & 40.01 & 39.40 \\
Claude-3.5-Sonnet~\citep{anthropic_claude35sonnet} & \redmark & \redmark & 43.17 & 44.45 & 52.09 & 39.03 & 37.09 \\
o3-mini-high~\citep{openai_o3mini} & \greencheck & \redmark & 42.15 & 53.16 & 50.68 & 32.58 & 32.18 \\
Claude-3.7-Sonnet~\citep{anthropic_claude37sonnet} & \redmark & \redmark & 41.21 & 34.13 & 59.77 & 36.80 & 34.15 \\
o3-mini-medium~\citep{openai_o3mini} & \greencheck & \redmark & 39.25 & 52.17 & 42.66 & 32.31 & 29.84 \\
Doubao-1.5-Pro-32k~\citep{bytedance_doubao15pro32k} & \redmark & \redmark & 38.88 & 42.80 & 46.81 & 33.63 & 32.28 \\
DeepSeek-V3-0324~\citep{deepseek_v3_0324} & \greencheck & \greencheck & 37.31 & 45.34 & 46.18 & 27.70 & 30.02 \\
GPT-4o-1120~\citep{openai_gpt4o1120} & \redmark & \redmark & 35.83 & 39.28 & 31.26 & 36.69 & 36.12 \\
Gemini-2.0-Flash~\citep{google_gemini20flash} & \redmark & \redmark & 35.61 & 48.56 & 39.24 & 26.71 & 27.92 \\
Qwen3-235B-A22B~\citep{qwen3_235b} & \greencheck & \greencheck & 34.88 & 42.47 & 39.42 & 28.03 & 29.59 \\
Llama4-Maverick~\cite{llama4Maverick} & \redmark & \greencheck & 34.50 & 44.96 & 32.46 & 30.77 & 29.83 \\
DeepSeek-V3-1226~\citep{deepseek_v3_1226} & \greencheck & \greencheck & 33.16 & 37.23 & 37.08 & 28.71 & 29.63 \\
GPT-4.1-mini-0414~\citep{openai_gpt41mini0414} & \redmark & \redmark & 31.23 & 40.23 & 30.17 & 26.39 & 28.13 \\
Qwen3-32B~\cite{qwen3_30b} & \greencheck & \greencheck & 30.96 & 38.35 & 27.82 & 30.92 & 26.74 \\
Qwen2.5-Max~\citep{qwen_max} & \redmark & \redmark & 30.41 & 39.77 & 31.90 & 26.76 & 23.22 \\
Qwen3-30B-A3B~\cite{qwen3_30b} & \greencheck & \greencheck & 29.23 & 48.91 & 17.53 & 26.36 & 24.10 \\
Qwen2.5-72B-Instruct~\citep{qwen25_72binstruct} & \redmark & \greencheck & 29.21 & 38.38 & 30.41 & 24.06 & 23.97 \\
Qwen3-14B~\cite{qwen3_14b} & \greencheck & \greencheck & 28.27 & 42.15 & 21.26 & 24.84 & 24.83 \\
Llama4-Scout~\cite{llama4Scout} & \redmark & \greencheck & 27.27 & 43.36 & 17.96 & 23.92 & 23.84 \\
Qwen3-8B~\cite{qwen3_8b} & \greencheck & \greencheck & 27.12 & 45.69 & 17.38 & 22.36 & 23.05 \\
GLM-Z1-Air~\citep{glm_zero_air0414} & \greencheck & \greencheck & 25.84 & 35.75 & 22.49 & 24.36 & 20.76 \\
Qwen3-4B~\cite{qwen3_4b} & \greencheck & \greencheck & 25.52 & 46.77 & 16.62 & 17.62 & 21.07 \\
\bottomrule
\end{tabular}
}
\end{table}

\section{Experiments}
\label{sec: experiments}

\subsection{Experiment Setup}
\label{subsec: Experiment Setup}
\paragraph{\textbf{Prompting Setting}}
\label{par: prompting strategy}

All models are evaluated under a zero-shot prompting strategy. 
Task instructions and dialogue scenarios are specified using carefully designed prompt templates tailored to each task category, as detailed in Appendix~\ref{appendix: task categories}.

\paragraph{\textbf{Evaluation Metrics}}
\label{par: evaluation metrics}

We adopt the LLM-as-a-judge paradigm to evaluate model outputs. A task-specific checklist is constructed to define explicit assessment criteria, based on which the judge model scores each predicted response. This approach is well-suited to our setting, as each question is associated with a definitive reference answer derived from standardized play-by-play records, and the evaluation criteria are designed to be objective and transparent. Given the fixed task structure and limited ambiguity, mainstream LLMs such as \texttt{DeepSeek-V3-0324} are able to perform reliable and consistent judgments across all experiments, including those presented in Section~\ref{sec:discussion}. Details of the evaluation prompt and scoring format are provided in Appendix~\ref{appendix:metrics}.

\subsection{Experiment Results}
\label{subsec: Experiment Setup}

Table~\ref{tab:main_result} presents the evaluation results of various representative models on \ourbenchmark{}. Based on these results, we summarize the following key observations.

\paragraph{\textbf{LLMs Struggle in Complex Multi-turn Dialogues.}} Even top models achieve around 70 points, with performance decreasing as dialogue turns increase, shown in Section~\ref{subsec:rq1}, highlighting limitations in handling extended multi-turn interactions. Lower scores on IF, IR, and particularly TS tasks further underscore deficiencies in cross-turn context management and interactive scenarios. In addition, model size affects performance, with smaller models generally exhibiting lower scores (see Appendix~\ref{appendix:model_size_analysis}).

\paragraph{\textbf{Closed-Source Models Lead in Complex Multi-Turn Scenarios.}} In challenging multi-turn dialogue tasks, closed-source models consistently outperform open-source counterparts. For example, Google’s Gemini-2.5-Pro achieves a 72.44 overall on MARS-Bench under complex contextual and reasoning requirements, while the top open-source DeepSeek-R1 reaches just 45.42. Open-source models—though flexible—often lack the scale and targeted optimization needed to excel in intricate information reasoning and task-switching.

\paragraph{\textbf{Reasoning Models demonstrate greater performance.}}

Models equipped with chain-of-thought reasoning tend to engage more deliberate, System 2-style inference and decision-making processes. As a result, they exhibit higher consistency and correctness across multi-turn dialogue tasks. In contrast, models that rely on System 1-style heuristic generation are more susceptible to variations in task complexity and context, leading to comparatively weaker overall performance. For example, DeepSeek-R1 achieves an overall MARS-Bench score of 45.42, outperforming DeepSeek-V3 (37.31) by 8.11 points, and even scores over 12 points higher on information reasoning (40.01 vs. 27.70). We also observe that applying CoT prompting to non-reasoning models improves performance for most non-reasoning models(see Appendix~\ref{appdendix:CoT}).

\paragraph{\textbf{Models perform worse on the Instruction Following task.}}

Both reasoning-enhanced and standard models demonstrate relatively poor performance on the instruction following task. Analysis reveals that current models struggle to track turn-level structures as required by system prompts. In particular, they often fail to produce the correct output in the specified dialogue turn, suggesting limitations in their ability to align generation behavior with round-dependent instructions.

\begin{figure*}[t]
    \centering
    \begin{subfigure}[t]{0.48\textwidth}
        \centering
        \includegraphics[width=\textwidth]{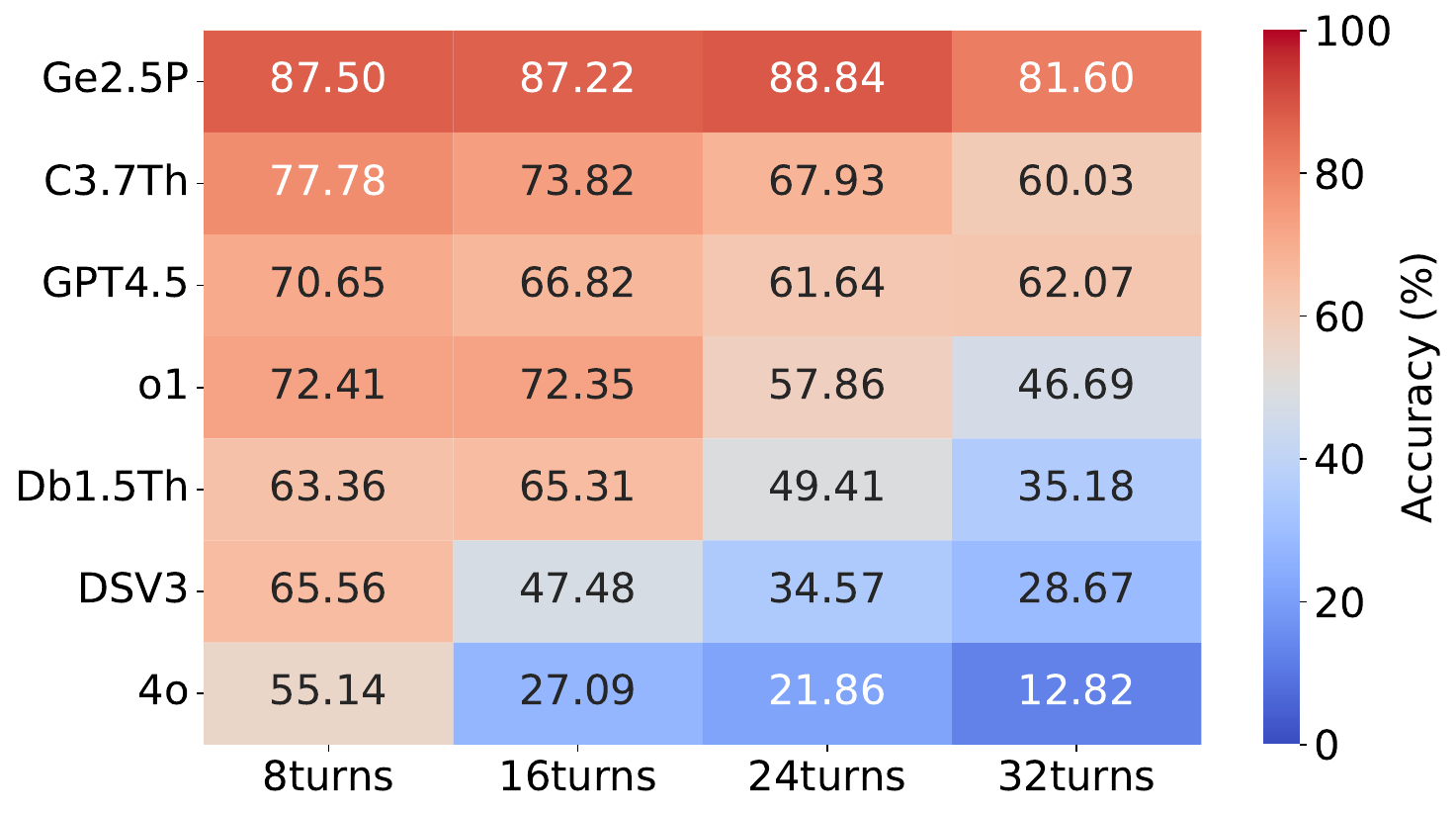}
        \caption{Accuracy trends in the Context Retrieval task}
        \label{fig:heatmap_main_cr}
    \end{subfigure}
    \hfill
    \begin{subfigure}[t]{0.48\textwidth}
        \centering
        \includegraphics[width=\textwidth]{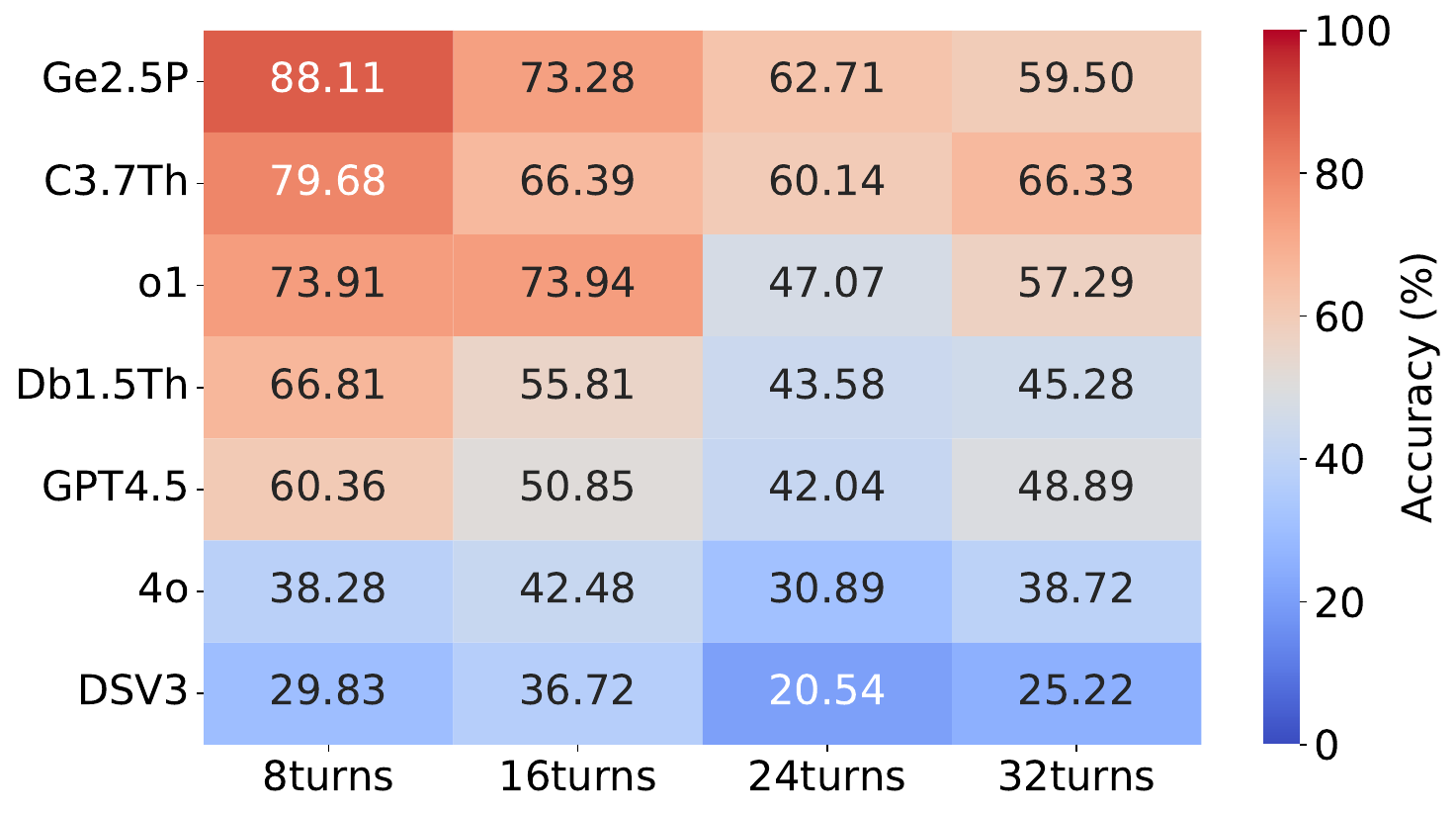}
        \caption{Accuracy trends in the Information Reasoning task}
        \label{fig:heatmap_main_ir}
    \end{subfigure}
    \caption{\textbf{Accuracy degradation in CR and IR tasks with increasing input turns.} Subfigures (a) and (b) show results for the Context Retrieval (CR) and Information Reasoning (IR) tasks, respectively. \texttt{Db1.5Th} refers to \texttt{Doubao-1.5-Pro-Thinking}, \texttt{Ge2.5P} to \texttt{Gemini-2.5-Pro}, and \texttt{C3.7Th} to \texttt{Claude-3.7-Sonnet-Thinking}. See Appendix~\ref{appendix:model_names} for the full list of model names and abbreviations.}
    \label{fig:heatmap_main}
\end{figure*}

\section{Discussion}
\label{sec:discussion}


We organize our discussion around the following research questions:
(1) whether increasing the number of dialogue turns harms model performance;
(2) whether cross-turn context hinders reasoning; and
(3) whether interactive multi-turn generation degrades LLMs performance.
\subsection{LLMs Struggle with More Turns}
\label{subsec:rq1}

\paragraph{\textbf{Increasing Turns Leads to Lower Accuracy.}} 
As an additional analysis setting, we break down the main experimental results by interaction rounds to examine how performance evolves overturns.
As shown in Figure~\ref{fig:heatmap_main_cr} and Figure~\ref{fig:heatmap_main_ir}, model accuracy in context retrieval and information reasoning tasks tends to decline in the later stages of multi-turn interactions.

\begin{figure}[t]
    \centering
    \includegraphics[width=0.55\textwidth]{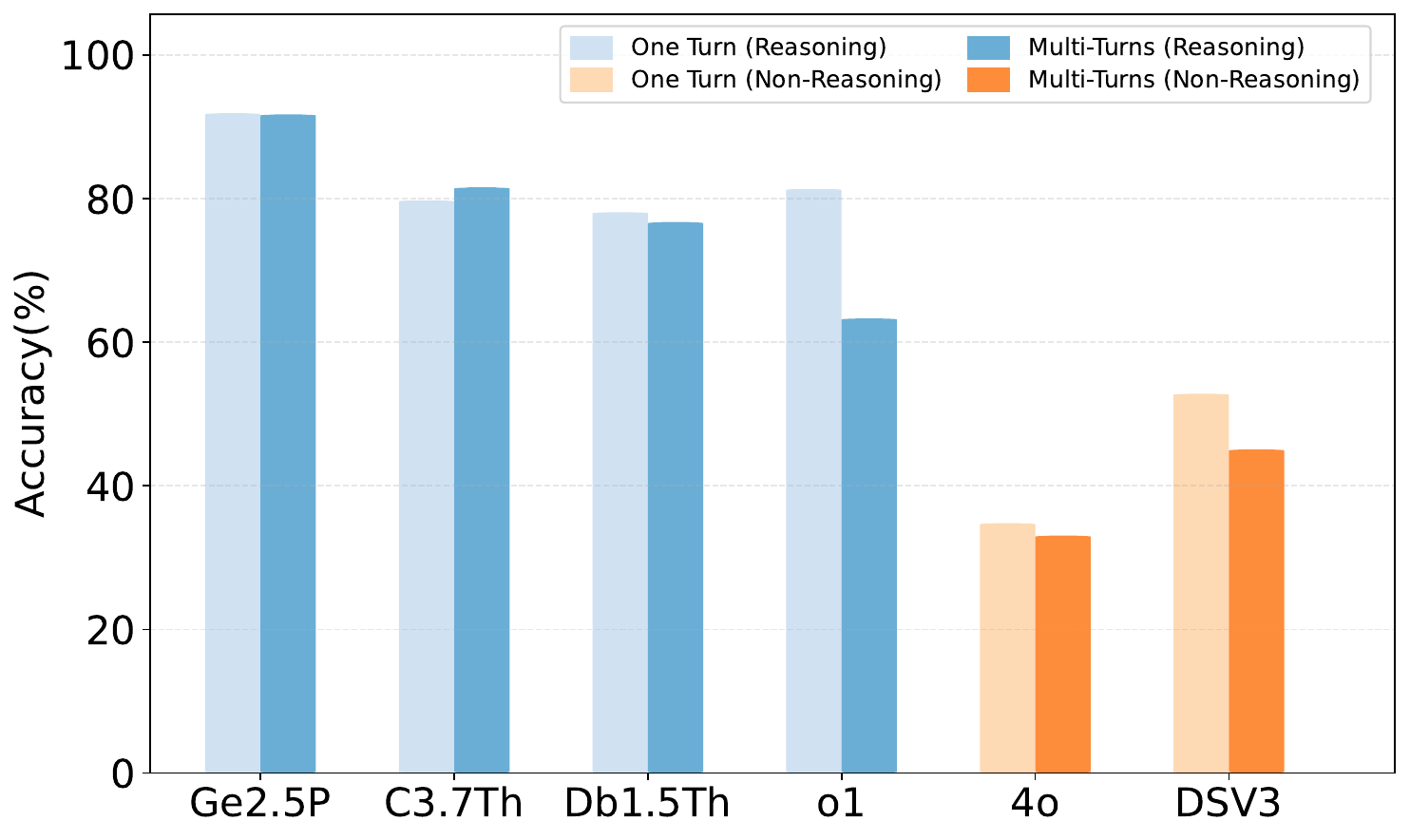}
    \caption{\textbf{Ablation on input format: multi-turn vs. single-turn.} Most models perform worse with multi-turn inputs, suggesting that input fragmentation hinders reasoning. Top models like \texttt{Gemini-2.5-Pro} and \texttt{Claude-3.7-Sonnet-Thinking} remain robust. Additional results are provided in Appendix~\ref{appendix:Details of Ablation Results}.}
    \label{fig:bar_ablation_turn_c}
\end{figure}

\paragraph{\textbf{Special Tokens Consume Attention in Multi-turn Contexts.}}
Motivated by the observed degradation in later turns, we further investigate whether the number of interaction rounds, independent of information content, contributes to the performance drop. We conduct an ablation study using identical play-by-play records presented in two formats: a 20-turn dialogue and a single-turn concatenation.
As shown in Figure~\ref{fig:bar_ablation_turn_c}, most models perform worse in the multi-turn setting, except for top models like \texttt{Gemini-2.5-Pro}.

To further illustrate this effect, we use a mechanistic interpretability approach to visualize attention in Qwen2.5-7B-Instruct, as shown in Figure~\ref{fig:aggregation_attention}. The multi-turn format introduces more special tokens, which absorb a notable portion of attention (e.g., ``\texttt{<|im\_end|>}''), reducing attention efficiency and contributing to performance degradation. Further implementation details are provided in Appendix~\ref{appendix:attention}.

\subsection{LLMs Fail Cross-Turn Context}
\label{subsec: rq2}

\begin{figure*}[t]
    \centering
    \begin{subfigure}[t]{0.48\textwidth}
        \centering
        \includegraphics[width=\textwidth]{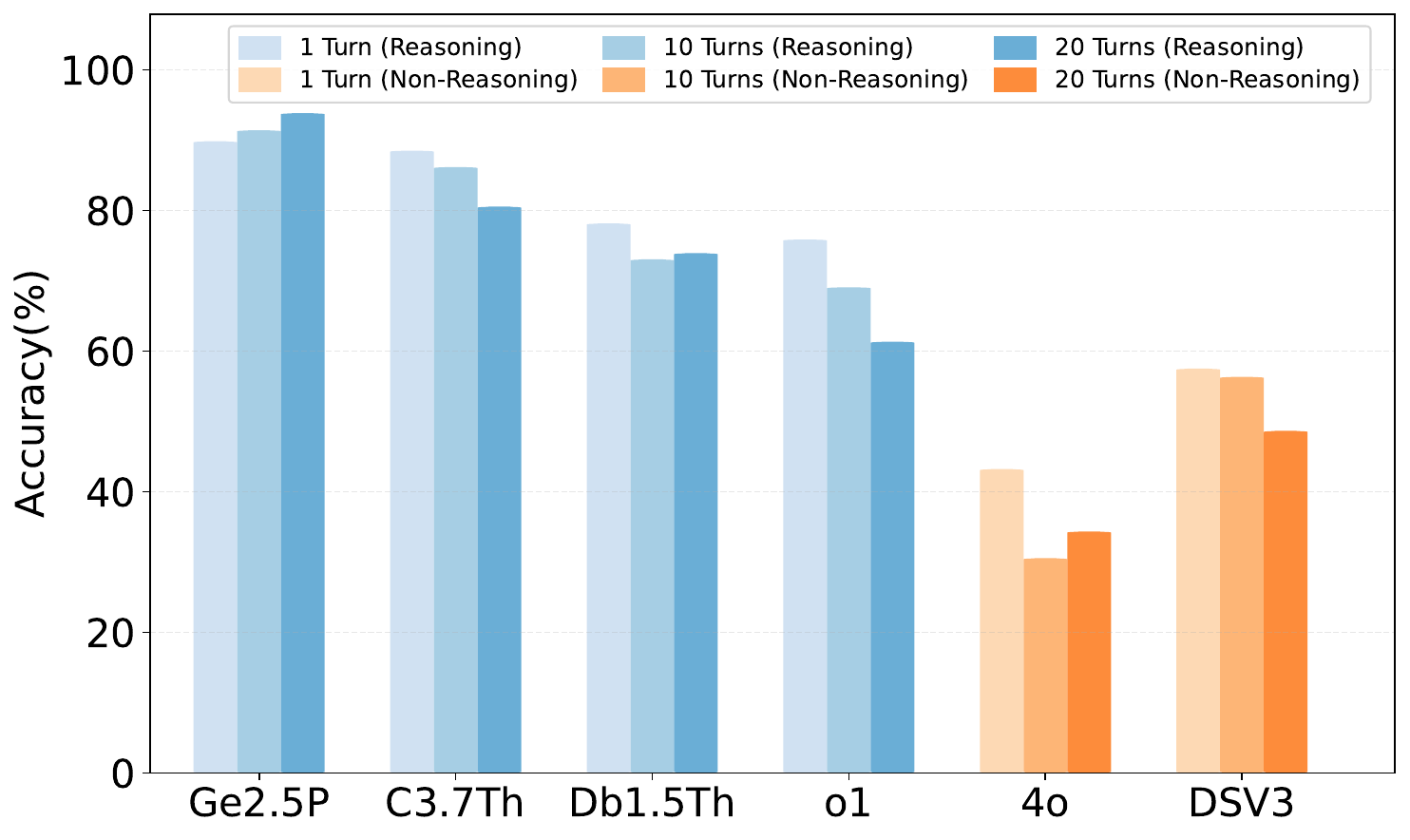}
        \caption{Performance under different turn lengths with the same content ($x = 1, 10, 20$).}
        \label{fig:aggregation_accuracy}
    \end{subfigure}
    \hfill
    \begin{subfigure}[t]{0.48\textwidth}
        \centering
        \includegraphics[width=\textwidth]{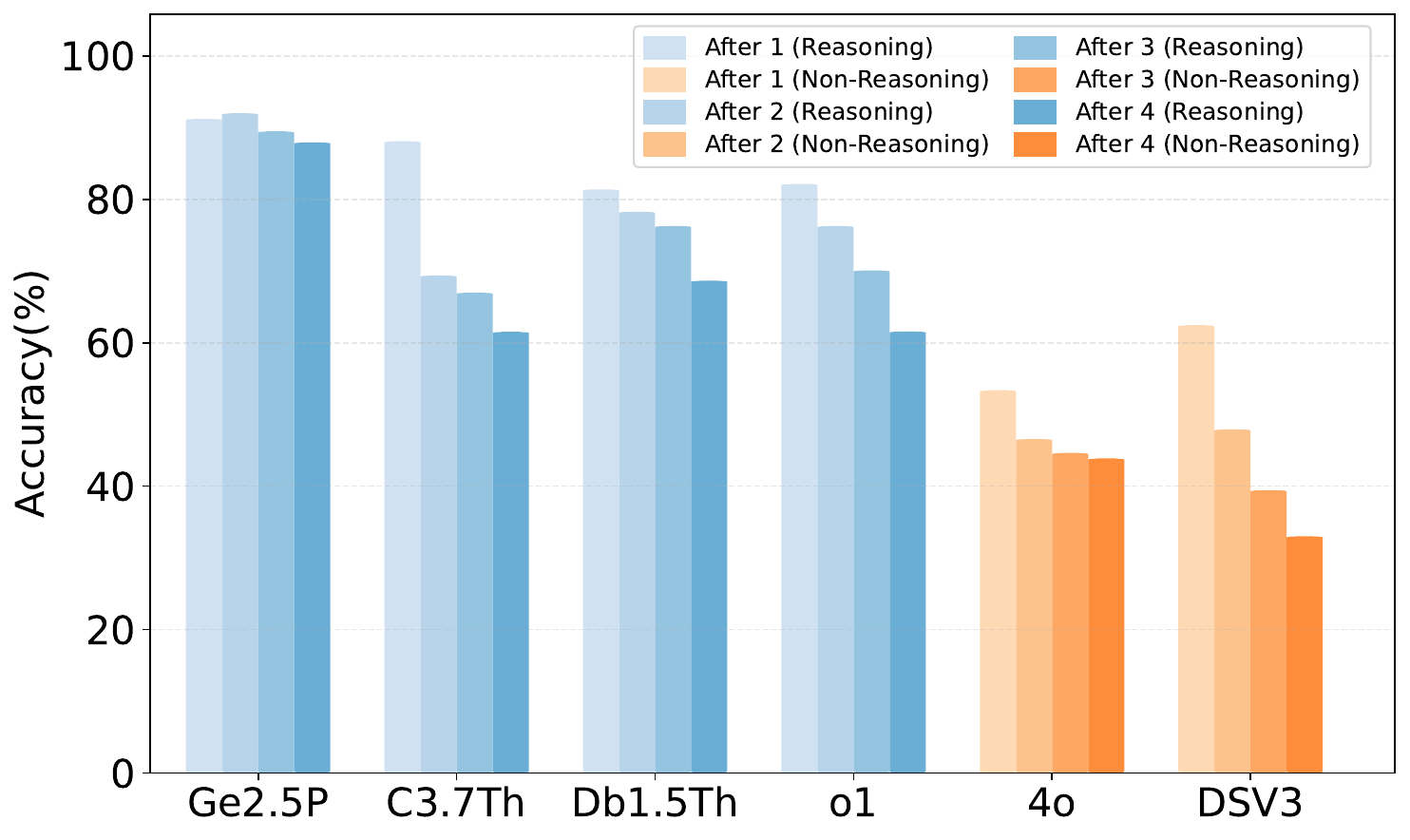}
        \caption{Performance under increasing retrieval distances.}
        \label{fig:retrieval_accuracy}
    \end{subfigure}
    \caption{\textbf{Ablation study on cross-turn context.} (a) Splitting identical content across more dialogue turns ($x = 1, 10, 20$) results in reduced performance. (b) Accuracy decreases as models are required to recall first-quarter information after each section (``After 1'' indicates the question is posed immediately following Q1).}
   \label{fig:ablation_combined}
\end{figure*}
\begin{figure}[t]
    \centering
    \begin{subfigure}[t]{0.48\linewidth}
        \centering
        \includegraphics[width=\linewidth]{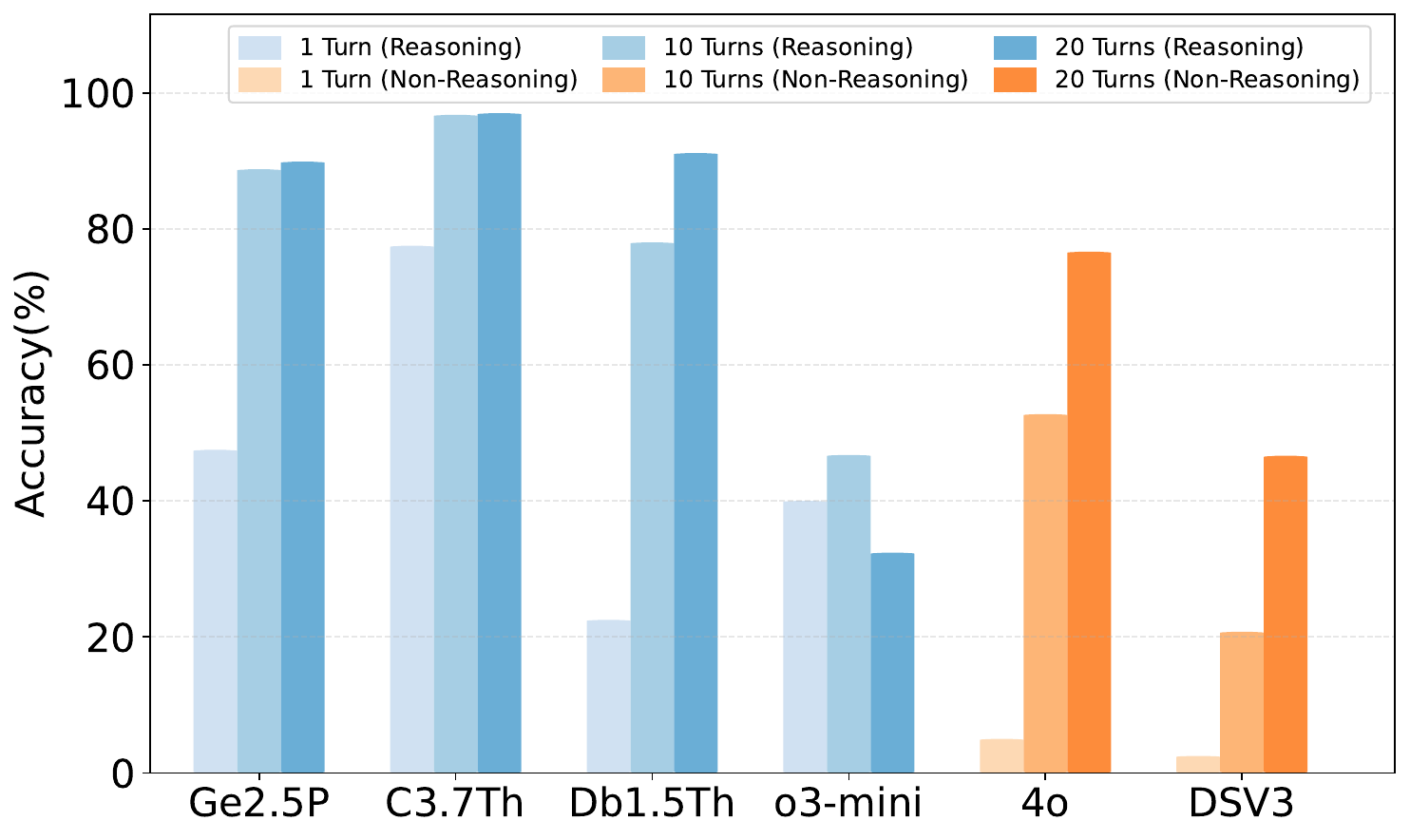}
        \caption{\textbf{Effect on Overall Accuracy.} Performance on score updates under different interaction turns (1, 10, 20).}
        \label{fig:interactive_accuracy}
    \end{subfigure}
    \hfill
    \begin{subfigure}[t]{0.48\linewidth}
        \centering
        \includegraphics[width=\linewidth]{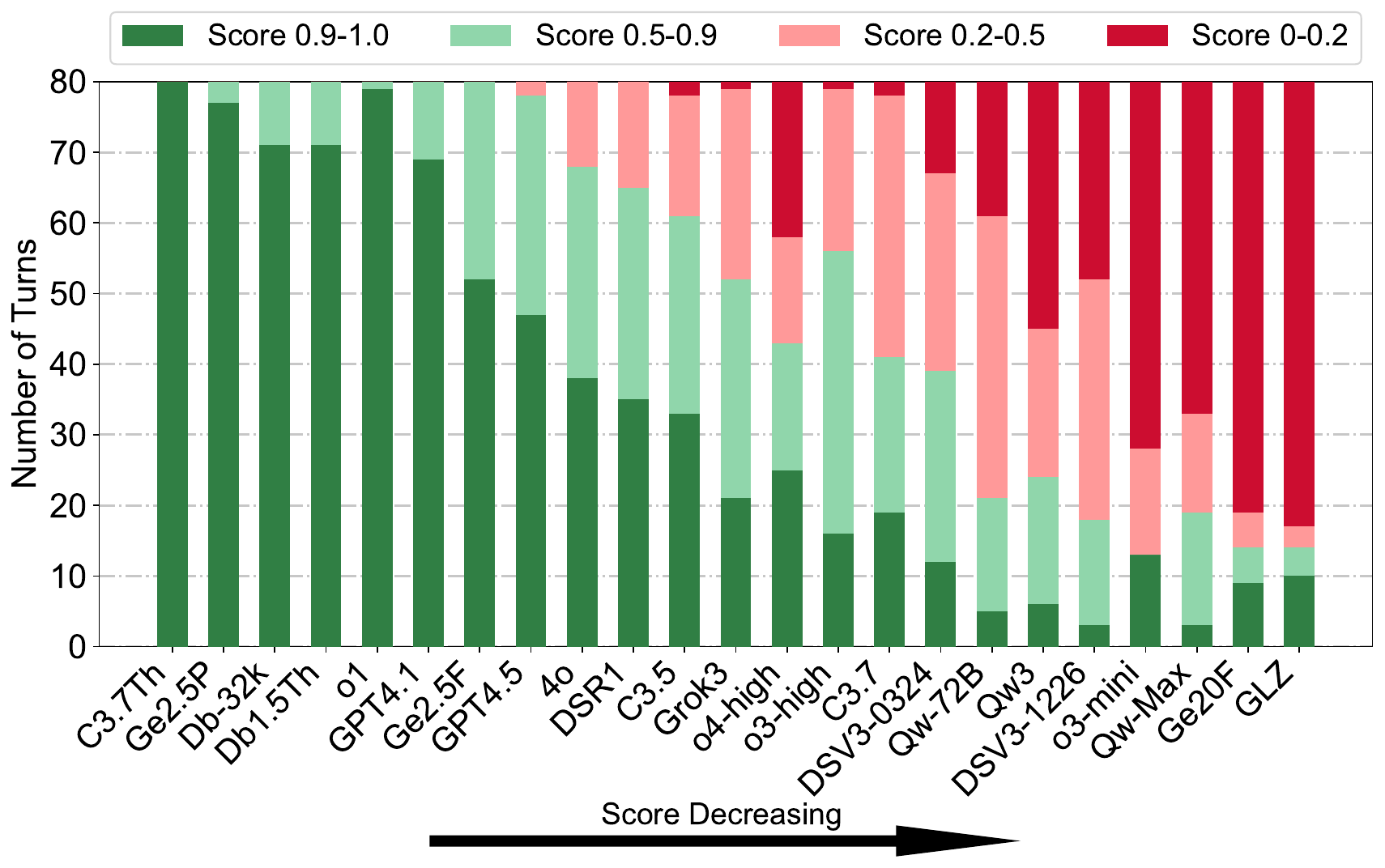}
        \caption{\textbf{Error Accumulation Effect.} Early mistakes accumulate in multi-turn settings, degrading final accuracy.}
        \label{fig:good_poor}
    \end{subfigure}
    \caption{\textbf{Ablation study on interaction turns.} (a) Longer turns to improve performance; (b) Early errors hurt in multi-turn settings.}
    \label{fig:turn_ablation_combined}
\end{figure}
\paragraph{\textbf{Distributed Information Reduces Focus on Relevant Context}}
We split an equal amount of play-by-play game records into different turn lengths ($x = 1, 10, 20$) and evaluated models on quarter-level statistical questions. As shown in Figure~\ref{fig:aggregation_accuracy}, accuracy generally declines with more turns, except for \texttt{Gemini-2.5-Pro}, which maintains or slightly improves its performance. 

\begin{figure*}[t]
\centering
\includegraphics[width=\textwidth]{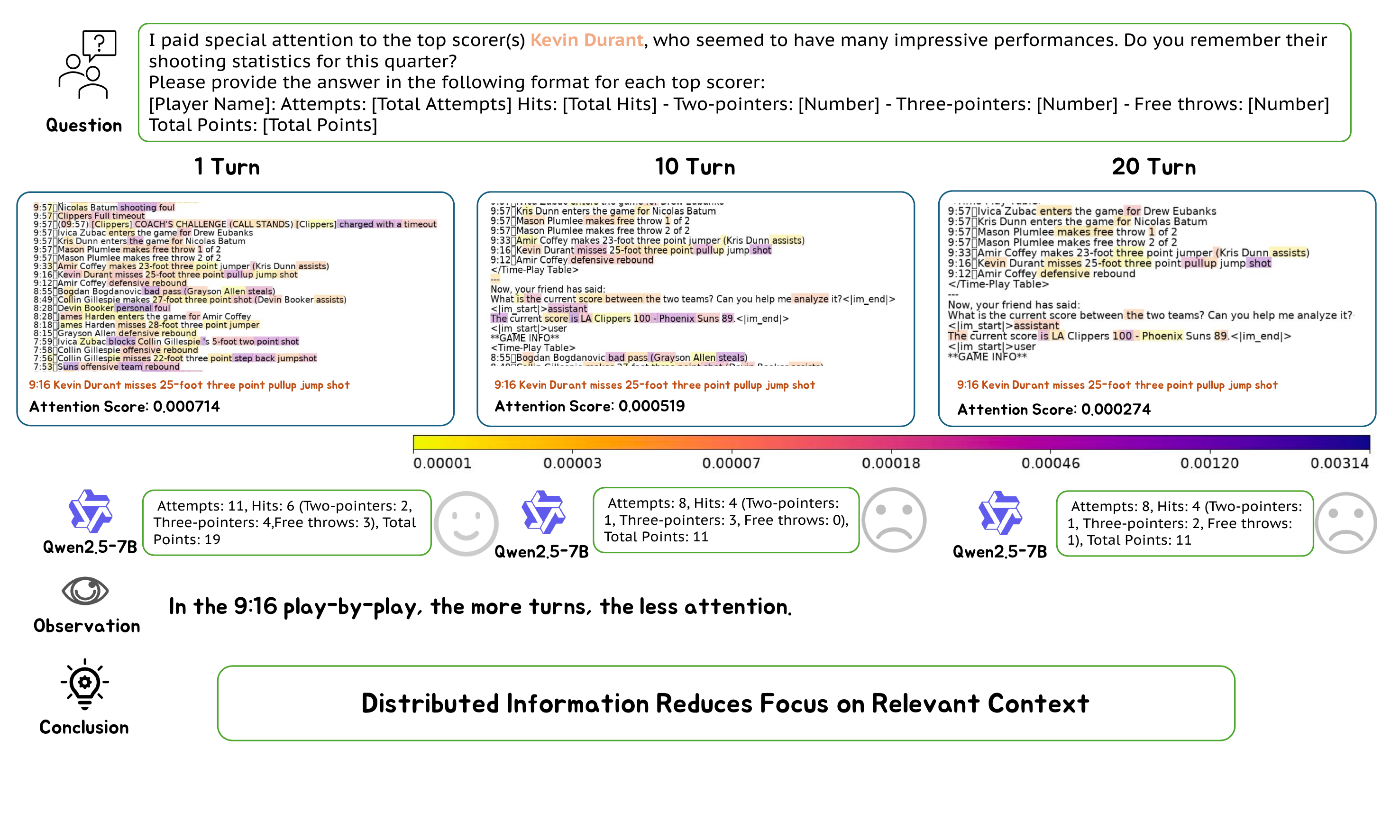}
\caption{\textbf{Attention visualization in \texttt{Qwen2.5-7B-Instruct}.} Attention to key content drops from 0.000714 (1 turn) to 0.000274 (20 turns), suggesting degraded focus in longer dialogues, indicating a 2.6× decay in attention to key content.}
\label{fig:turn_attention_qwen}
\end{figure*}

To gain further insight into the model's attention patterns, we visualize attention in \texttt{Qwen2.5-7B-Instruct}. Figure~\ref{fig:turn_attention_qwen} shows how the model's attention to key information changes across settings, revealing that longer dialogues impair its ability to attend effectively to relevant content. Further implementation details are provided in Appendix~\ref{appendix:attention}.

\paragraph{\textbf{Distant Context Weakens Retrieval Accuracy}}

We evaluate the models' long-range recall ability by asking questions about the first quarter after each subsequent quarter. 
As shown in Figure~\ref{fig:retrieval_accuracy}, accuracy declines as the retrieval distance increases, with substantial variation across models. 
While \texttt{Gemini-2.5-Pro} maintains consistent accuracy across settings, most models (e.g., \texttt{DeepSeek-V3-0324}) exhibit substantial performance degradation with increasing retrieval distance.

\subsection{LLMs Underperform in Interaction}
\label{subsec: rq3}

\paragraph{\textbf{Error Accumulation Degrades Interactive Performance}}
We partition the play-by-play records of each quarter into turn settings of varying lengths ($x = 1, 10, 20$), where the model predicts the current score at each turn. The total information remains fixed, but more turns reduce per-turn complexity. 
As shown in Figure~\ref{fig:interactive_accuracy}, most models improve with more interaction turns, but \texttt{o3-mini-high} and \texttt{o3-mini-medium} exhibit inconsistent trends. 
To better understand these inconsistencies and how performance evolves overturns, we conduct a turn-level analysis under the 20-turn setting. As illustrated in Figure~\ref{fig:good_poor}, strong models maintain stable accuracy across turns, whereas models with solid reasoning ability, such as \texttt{o3-mini-high} and \texttt{o3-mini-medium}, are hindered by early errors that accumulate and degrade overall performance.

\begin{figure*}[t]
    \centering
    \includegraphics[width=\textwidth]{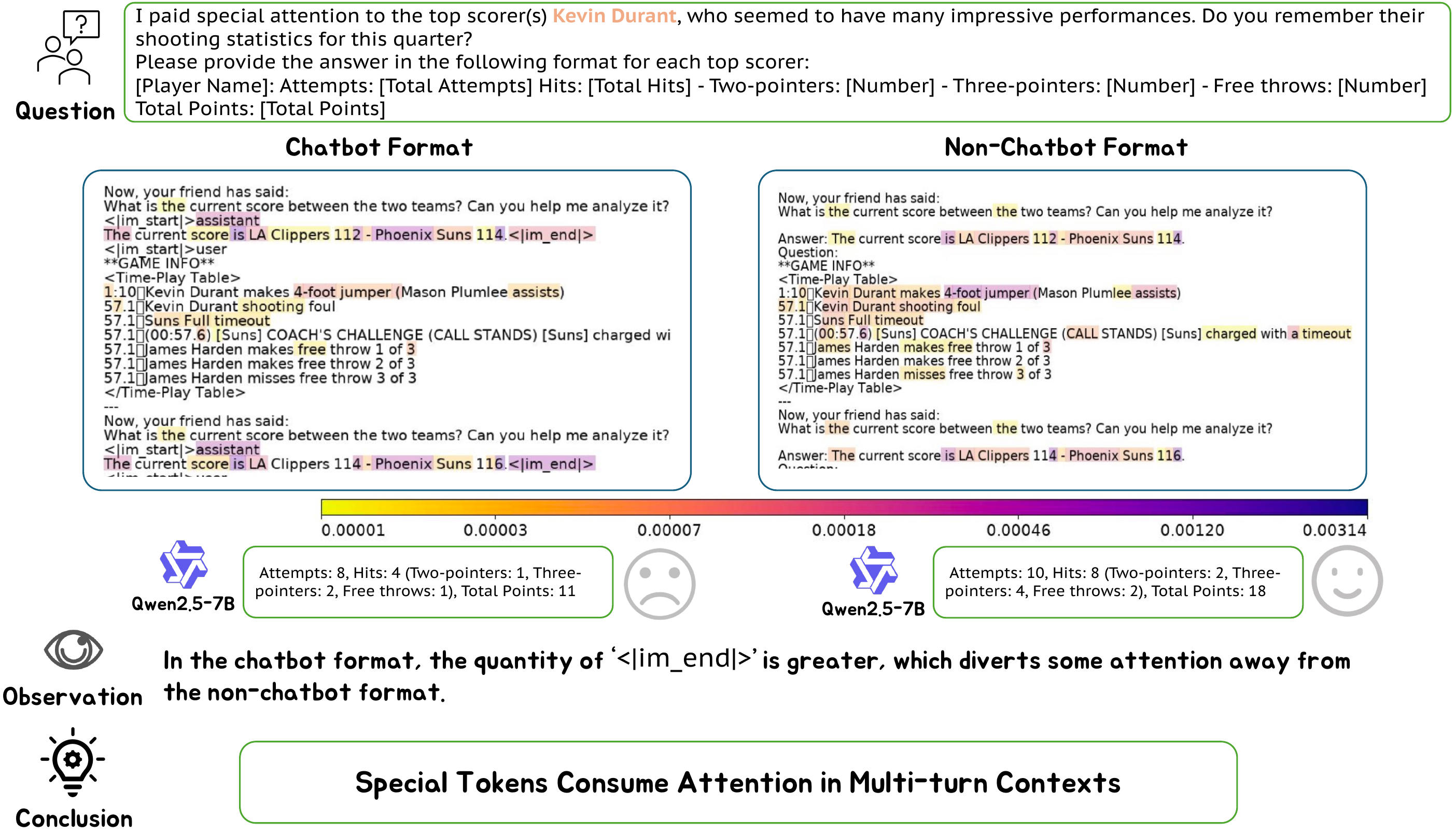}
    \caption{\textbf{Attention visualization in \texttt{Qwen2.5-7B-Instruct}.}
Multi-turn inputs introduce more special tokens, which absorb a substantial portion of attention (e.g., ``\texttt{<|im\_end|>}''), potentially reducing attention efficiency.}
    \label{fig:aggregation_attention}
\end{figure*}

\section{Conclusion}
We introduce MARS-Bench, a benchmark constructed from real-world play-by-play sports data to evaluate LLMs in complex multi-turn dialogue settings. It defines four task types: instruction following, context retrieval, information reasoning, and task switching, enabling systematic and fine-grained assessment of long-context dialogue capabilities. Experimental results suggest that models employing explicit reasoning strategies tend to perform more consistently, although all models exhibit persistent challenges in instruction alignment and context retention. Further analysis indicates that dialogue depth, input structure, and the accumulation of errors across turns can significantly affect model performance. MARS-Bench provides a realistic and focused benchmark for advancing research on multi-turn dialogue understanding.

\section{Related Work}

\paragraph{\textbf{Multi-turn Dialogue Evaluation Benchmarks}}

Multi-turn dialogue capability is a key research area for large language models. Early benchmarks such as MT-Bench~\citep{MT-Bench} and MT-Eval~\citep{MT-Eval} primarily focused on short-turn dialogues and basic instruction following. As model capabilities have improved, these evaluations have become less effective at distinguishing model performance. MT-Bench++~\citep{Parrot} expanded the dialogue length to eight turns, while MultiChallenge~\citep{MultiChallenge} introduced various task forms with five-turn dialogues. MINT~\citep{MINT} added user feedback and tool usage, increasing the interaction complexity. Despite progress in task coverage and interactivity, most benchmarks still rely on synthetic data and static dialogue settings, limiting their ability to reflect the dynamic evolution of context across turns. There remains a lack of systematic evaluation for key capabilities such as cross-turn reasoning and information tracking.

\paragraph{\textbf{Agent-Centric Evaluation of Interactive and Ultra-Long Multi-Turn Dialogues}}

Recent studies have shifted toward longer and more complex evaluation frameworks to better simulate real-world interactions. OpenAI introduced AlreadySaidThat and TrackTheState~\citep{OpenaiEvals} to assess historical consistency and cross-turn reasoning. LongMemEval~\citep{LongMemEval} tests long-term information retention, while LTM-Benchmark~\citep{LTM-Bench} evaluates task-switching in long-form dialogues. AgentBench~\citep{AgentBench} and RealWebAssist~\citep{RealWebAssist} expand evaluations to complex environments like database and web tasks. However, a unified framework that covers Interactive Multi-turn (IMT), Cross-turn Tasks (CTT), and Ultra Multi-turn (UMT) dialogues, addressing long-range dependency, task switching, and real-data interaction, is still lacking.

\section*{Limitations}

While MARS-Bench offers a structured and realistic setting for evaluating multi-turn dialogue, several limitations remain:
\begin{itemize}[leftmargin=*, itemsep=0pt]
\item \textbf{Domain specificity}: The focus on sports scenarios may limit generalizability to open-domain or everyday dialogues.
\item \textbf{Modality constraints}: The benchmark is limited to text and does not include multimodal inputs such as vision or speech.
\item \textbf{Evaluation method}: Checklist-based automatic scoring with LLM judges may miss subtle issues in coherence, style, or pragmatics.
\end{itemize}
Future work may explore broader domains, expand dataset scale, incorporate multimodal inputs, and include human evaluation to improve generalizability and coverage.

\clearpage

\bibliographystyle{plainnat}
\bibliography{main}

\clearpage

\beginappendix

\section{Model Abbreviations and Full Names}
\label{appendix:model_names}
A complete mapping between model abbreviations and their full names is presented in Table~\ref{tab:model_abbreviations}.

\begin{table}[h]
\renewcommand{\arraystretch}{1.1}
\centering
\caption{List of model abbreviations and their corresponding full names.}
\begin{tabular}{ll}
\toprule
\textbf{Abbreviation} & \textbf{Full Model Name} \\
\midrule
4o          & GPT-4o-1120 \\
C3.5        & Claude-3.5-Sonnet \\
C3.7        & Claude-3.7-Sonnet \\
C3.7Th      & Claude-3.7-Sonnet-Thinking \\
Db-32k      & Doubao-1.5-Pro-32k \\
Db1.5Th     & Doubao-1.5-Pro-Thinking \\
DSR1        & DeepSeek-R1 \\
DSV3        & DeepSeek-V3-0324 \\
DSV3-1226   & DeepSeek-V3-1226 \\
Ge20F       & Gemini-2.0-Flash \\
Ge2.5F      & Gemini-2.5-Flash  \\
Ge2.5P      & Gemini-2.5-Pro \\
GLZ         & GLM-Z1-Air \\
GPT-4.5     & GPT-4.5-Preview \\
GPT4.1      & GPT-4.1-mini-0414 \\
Grok3       & Grok3 \\
Ll4-M       & Llama4-Maverick\\
Ll4-S       & Llama4-Scout\\
o1          & o1-1217 \\
o3-high     & o3-mini-high \\
o3-mini     & o3-mini-medium \\
o4-high     & o4-mini-0416 \\
Qw3         & Qwen3-235B-A22B \\
Qw3-32      & Qwen3-32B \\
Qw3-30      & Qwen3-30B-A3B \\
Qw3-14      & Qwen3-14B \\
Qw3-8       & Qwen3-8B \\
Qw3-4       & Qwen3-4B \\
Qw-72B      & Qwen2.5-72B-Instruct \\
Qw-Max      & Qwen2.5-Max \\
\bottomrule
\end{tabular}%
\label{tab:model_abbreviations}
\end{table}

\section{Additional Experiment Data}
\label{appendix: results with other type}
In the TS task, models handle both context-relevant sports tasks and unrelated math problems. While the Table~\ref{tab:main_result} presents only the performance on sports tasks, the full results including math problems are shown in Table~\ref{tab:result_othertasks}.

\begin{table*}[h!]
\renewcommand{\arraystretch}{1}
\setlength{\extrarowheight}{3pt}
\centering\
\vskip 0.15in
\caption{Performance of different models on various task types. Additional results are provided for unrelated subtasks (e.g., mathematics) within the Task Switching (TS) category.}
\small
\resizebox{\textwidth}{!}{
\begin{tabular}{lccccccc}
\toprule
\multirow{2}{*}{\textbf{Model}} & \multirow{2}{*}{\textbf{Overall}} & \multirow{2}{*}{\textbf{IF}} & \multirow{2}{*}{\textbf{CR}} & \multirow{2}{*}{\textbf{IR}} & \multicolumn{2}{c}{\textbf{TS}} \\
 \cmidrule(lr){6-7}
& & & \textbf{} & \textbf{} & \textbf{Sports Games} & \textbf{Other Tasks} \\
\midrule
Gemini-2.5-Pro~\citep{google_gemini25pro0506} & \first{72.44} & \first{65.08} & \first{87.06} & \first{70.92} & \second{66.72} & \first{89.57} \\ 
Claude-3.7-Sonnet-Thinking~\citep{anthropic_claude37sonnet} & \second{62.29} & 43.28 & 71.51 & \second{66.98} & \first{67.38} & 86.96 \\ 
o1-1217~\citep{openai_o11217} & 59.62 & 53.09 & 64.48 & \third{62.63} & 58.28 & \third{87.55} \\ 
Gemini-2.5-Flash~\citep{google_gemini25flash} & 59.22 & 45.96 & \second{77.76} & 52.93 & \third{60.23} & 87.54 \\ 
GPT-4.5-Preview~\citep{openai_gpt45preview} & 53.33 & \third{55.52} & 66.65 & 50.43 & 40.74 & 65.80 \\ 
Doubao-1.5-Pro-Thinking~\citep{bytedance_doubao15prothinking} & 52.62 & 51.99 & 55.64 & 52.17 & 50.69 & \second{88.12} \\ 
Grok3~\citep{grok3} & 51.21 & \second{61.19} & \third{73.91} & 33.89 & 35.87 & 57.97 \\ 
o4-mini-0416~\citep{openai_o4mini} & 47.13 & 47.48 & 61.26 & 39.74 & 40.03 & 79.71 \\ 
DeepSeek-R1~\citep{deepseek_r1} & 45.42 & 53.04 & 49.23 & 40.01 & 39.40 & 85.80 \\ 
Claude-3.5-Sonnet~\citep{anthropic_claude35sonnet} & 43.17 & 44.45 & 52.09 & 39.03 & 37.09 & 87.25 \\ 
o3-mini-high~\citep{openai_o3mini} & 42.15 & 53.16 & 50.68 & 32.58 & 32.18 & 87.54 \\ 
Claude-3.7-Sonnet~\citep{anthropic_claude37sonnet} & 41.21 & 34.13 & 59.77 & 36.80 & 34.15 & 87.25 \\ 
o3-mini-medium~\citep{openai_o3mini} & 39.25 & 52.17 & 42.66 & 32.31 & 29.84 & 87.54 \\ 
Doubao-1.5-Pro-32k~\citep{bytedance_doubao15pro32k} & 38.88 & 42.80 & 46.81 & 33.63 & 32.28 & 72.46 \\ 
DeepSeek-V3-0324~\citep{deepseek_v3_0324} & 37.31 & 45.34 & 46.18 & 27.70 & 30.02 & 85.80 \\ 
GPT-4o-1120~\citep{openai_gpt4o1120} & 35.83 & 39.28 & 31.26 & 36.69 & 36.12 & 39.71 \\ 
Gemini-2.0-Flash~\citep{google_gemini20flash} & 35.61 & 48.56 & 39.24 & 26.71 & 27.92 & 68.41 \\ 
Qwen3-235B-A22B~\citep{qwen3_235b} & 34.88 & 42.47 & 39.42 & 28.03 & 29.59 & 79.71 \\ 
Llama4-Maverick~\cite{llama4Maverick} & 34.50 & 44.96 & 32.46 & 30.77 & 29.83 & 91.01\\
DeepSeek-V3-1226~\citep{deepseek_v3_1226} & 33.16 & 37.23 & 37.08 & 28.71 & 29.63 & 73.91 \\ 
GPT-4.1-mini-0414~\citep{openai_gpt41mini0414} & 31.23 & 40.23 & 30.17 & 26.39 & 28.13 & 75.36 \\
Qwen3-32B~\cite{qwen3_30b} & 30.96 & 38.35 & 27.82 & 30.92 & 26.74 &  90.43\\
Qwen2.5-Max~\citep{qwen_max} & 30.41 & 39.77 & 31.90 & 26.76 & 23.22 & 68.41 \\
Qwen3-30B-A3B~\cite{qwen3_30b} & 29.23 & 48.91 & 17.53 & 26.36 & 24.10 & 89.86\\
Qwen2.5-72B-Instruct~\citep{qwen25_72binstruct} & 29.21 & 38.38 & 30.41 & 24.06 & 23.97 & 76.81 \\
Qwen3-14B~\cite{qwen3_14b} & 28.27 & 42.15 & 21.26 & 24.84 & 24.83 & 88.12 \\
Llama4-Scout~\cite{llama4Scout} & 27.27 & 43.36 & 17.96 & 23.92 & 23.84 & 82.61\\
Qwen3-8B~\cite{qwen3_8b} & 27.12 & 45.69 & 17.38 & 22.36 & 23.05 & 87.83\\
GLM-Z1-Air~\citep{glm_zero_air0414} & 25.84 & 35.75 & 22.49 & 24.36 & 20.76 & 86.38 \\
Qwen3-4B~\cite{qwen3_4b} & 25.52 & 46.77 & 16.62 & 17.62 & 21.07 & 86.96\\
\bottomrule
\end{tabular}
}
\label{tab:result_othertasks}
\end{table*}

\section{Detailed Ablation Study Results}
\label{appendix:Details of Ablation Results}
This section presents detailed data for each ablation study, along with visualizations similar to those shown in the Section~\ref{sec:discussion}. Specific ablation results for different models are provided in Table~\ref{tab:ablation-results-table}.

\begin{table*}[h]
\renewcommand{\arraystretch}{1.1}
\setlength{\extrarowheight}{3pt} 
\centering
\caption{Detailed Ablation Data for Different Models.}
\resizebox{\textwidth}{!}{
\begin{tabular}{lccccccccccccc}
\toprule
\multirow{2}{*}{\textbf{Model}} & \multicolumn{3}{c}{\textbf{Ablation Interaction Turns}} & \multicolumn{3}{c}{\textbf{Ablation Turn Length}} & \multicolumn{2}{c}{\textbf{Ablation Input Format}} & \multicolumn{4}{c}{\textbf{Ablation Retrieval Distance}}\\
\cmidrule(lr){2-4} \cmidrule(lr){5-7} \cmidrule(lr){8-9} \cmidrule(lr){10-13} 
& \textbf{1 turn} & \textbf{10 turns} & \textbf{20 turns} & \textbf{1 turn} & \textbf{10 turns} & \textbf{20 turns} & \textbf{Long Text} & \textbf{Multi-turn} & \textbf{After 1 turn} & \textbf{After 2 turns} & \textbf{After 3 turns} & \textbf{After 4 turns} \\
\midrule
Gemini-2.5-Pro & 47.50 & 88.75 & 89.87 & 89.83 & 91.40 & 93.83 & 91.92 & 91.75 & 91.25 & 92.03 & 89.50 & 87.96 \\
Claude-3.7-Sonnet-Thinking & 77.50 & 96.75 & 97.00 & 88.50 & 86.17 & 80.54 & 79.78 & 81.58 & 88.11 & 69.39 & 67.00 & 61.53 \\
o1-1217 & 77.50 & 84.25 & 87.63 & 75.87 & 69.08 & 61.33 & 81.38 & 63.33 & 82.17 & 76.30 & 70.11 & 61.59 \\
Gemini-2.5-Flash & 10.00 & 77.00 & 91.09 & 78.63 & 74.96 & 75.82 & 76.96 & 77.33 & 82.20 & 77.96 & 77.41 & 81.03 \\
Doubao-1.5-Pro-Thinking & 22.50 & 78.00 & 91.13 & 78.15 & 73.04 & 73.92 & 78.13 & 76.75 & 81.41 & 78.29 & 76.30 & 68.67 \\
Grok3 & 22.50 & 44.25 & 57.25 & 64.58 & 64.63 & 64.13 & 63.25 & 65.83 & 74.02 & 64.09 & 59.50 & 63.74 \\
DeepSeek-R1 & 15.00 & 63.00 & 67.00 & 70.75 & 57.25 & 55.67 & 61.71 & 57.75 & 78.36 & 59.22 & 61.49 & 50.24 \\
o3-mini-high & 50.00 & 30.00 & 47.48 & 86.00 & 34.00 & 31.38 & 42.20 & 30.29 & 77.98 & 56.41 & 45.28 & 37.92 \\
o3-mini-medium & 40.00 & 46.75 & 32.38 & 77.25 & 25.96 & 23.33 & 32.04 & 23.63 & 64.22 & 41.21 & 36.93 & 32.33 \\
Claude-3.5-Sonnet & 25.00 & 58.75 & 63.88 & 67.42 & 54.17 & 48.67 & 50.48 & 49.32 & 54.33 & 56.83 & 43.35 & 42.82 \\
Qwen3-235B-A22B & 2.50 & 42.75 & 37.38 & 55.29 & 51.00 & 43.96 & 45.38 & 45.54 & 69.68 & 56.64 & 45.94 & 47.90 \\
Claude-3.7-Sonnet & 7.50 & 23.50 & 54.63 & 64.25 & 57.38 & 58.88 & 56.60 & 57.17 & 61.29 & 58.80 & 48.68 & 45.80 \\
GPT-4o-1120 & 5.00 & 52.75 & 76.63 & 43.21 & 30.54 & 34.33 & 34.79 & 33.07 & 53.38 & 46.58 & 44.64 & 43.87 \\
Doubao-1.5-Pro-32k & 35.00 & 72.50 & 93.13 & 51.96 & 44.42 & 42.92 & 49.08 & 44.67 & 55.77 & 41.14 & 36.56 & 41.99 \\
Llama4-Maverick & 5.00 & 53.25 & 39.75 & 43.21 & 37.08 & 31.21 & 39.00 & 31.50 & 51.69 & 46.68 & 35.98 & 38.01 \\
DeepSeek-V3-1226 & 2.50 & 30.25 & 33.88 & 58.75 & 46.71 & 41.44 & 51.96 & 44.88 & 67.72 & 47.55 & 40.05 & 36.28 \\
Gemini-2.0-Flash & 7.50 & 50.50 & 22.75 & 56.50 & 39.79 & 41.63 & 46.00 & 42.38 & 58.85 & 39.89 & 27.16 & 35.75 \\
DeepSeek-V3-0324 & 2.50 & 20.75 & 46.62 & 57.50 & 56.33 & 48.65 & 52.83 & 45.08 & 62.46 & 47.93 & 39.45 & 33.02 \\
Qwen3-32B & 5.00 & 55.25 & 65.25 & 51.79 & 23.38 & 27.88 & 33.88 & 33.04 & 59.82 & 44.04 & 43.63 & 32.22 \\
Qwen2.5-Max & 0.00 & 22.50 & 24.00 & 48.62 & 42.46 & 43.88 & 45.00 & 42.46 & 54.96 & 40.38 & 33.83 & 33.40 \\
Qwen3-30B-A3B & 0.00 & 45.00 & 49.63 & 46.83 & 23.17 & 30.83 & 32.08 & 23.58 & 60.41 & 50.47 & 35.64 & 23.87 \\
Qwen2.5-72B-Instruct & 7.50 & 21.50 & 39.63 & 37.62 & 34.92 & 36.25 & 35.58 & 37.58 & 51.24 & 45.96 & 40.10 & 37.26 \\
Qwen3-14B & 7.50 & 35.25 & 29.13 & 44.71 & 21.96 & 20.96 & 20.67 & 25.25 & 40.36 & 41.58 & 40.76 & 32.43 \\
Llama4-Scout & 2.50 & 42.25 & 16.50 & 33.58 & 32.67 & 35.79 & 39.32 & 33.67 & 44.56 & 40.10 & 26.12 & 23.25 \\
Qwen3-8B & 0.00 & 18.75 & 28.63 & 35.96 & 20.96 & 23.38 & 24.38 & 28.92 & 46.43 & 30.89 & 24.82 & 27.14 \\
Qwen3-4B & 2.50 & 6.25 & 10.50 & 29.83 & 16.79 & 19.63 & 21.00 & 20.83 & 33.42 & 19.13 & 21.60 & 21.55 \\
\bottomrule
\end{tabular}
}
\label{tab:ablation-results-table}
\end{table*}

\subsection{Ablation on Input Format}
\begin{figure*}[h!]
    \centering
    \includegraphics[width=\textwidth]{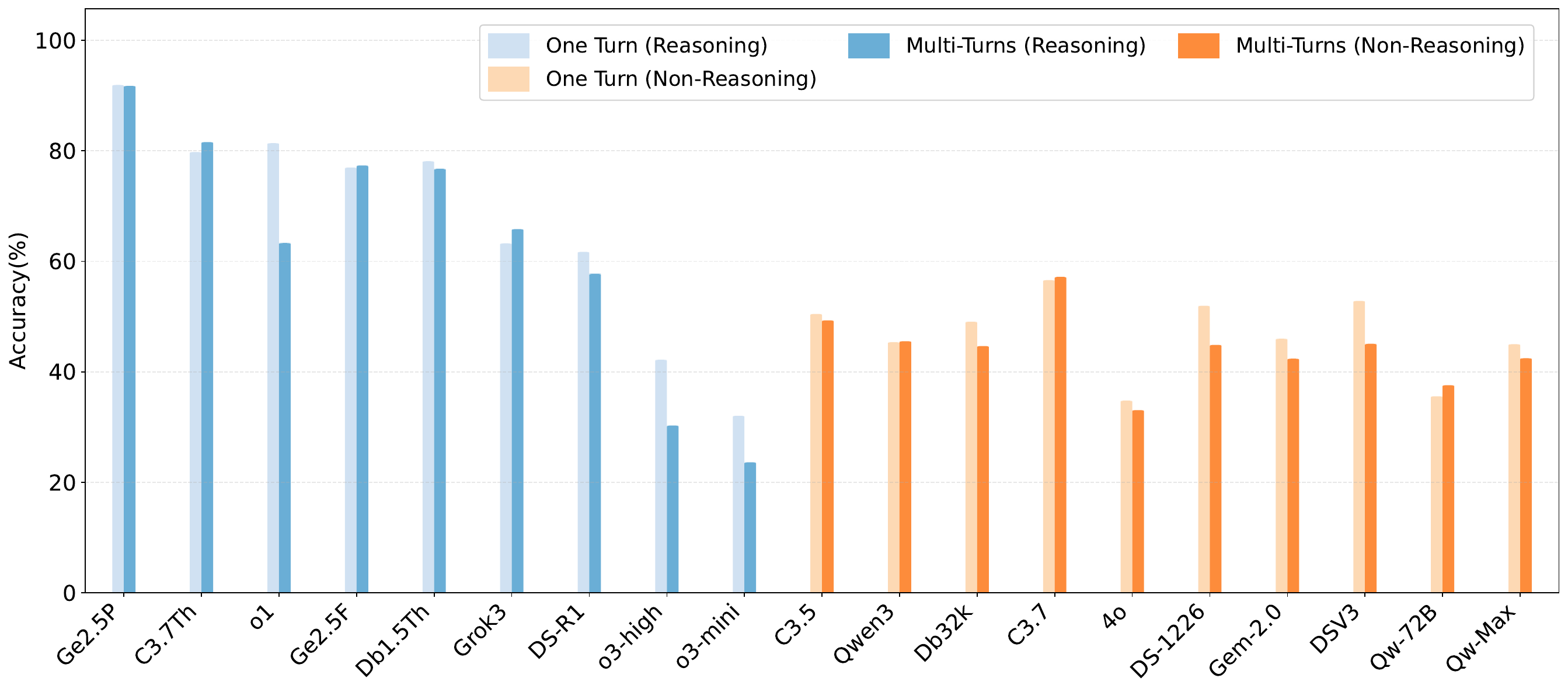}
    \caption{Detailed Ablation Results on Input Format}
    \label{fig:Ablation-Input Format}
\end{figure*}

In the Section~\ref{subsec:rq1}, we analyzed the impact of input formats on dialogue performance using several representative models. Here, we provide bar charts for the experimental results of all models (see Figure~\ref{fig:Ablation-Input Format}).

\subsection{Ablation on Cross-turn Context}
\begin{figure*}[h!]
    \centering
    \includegraphics[width=\textwidth]{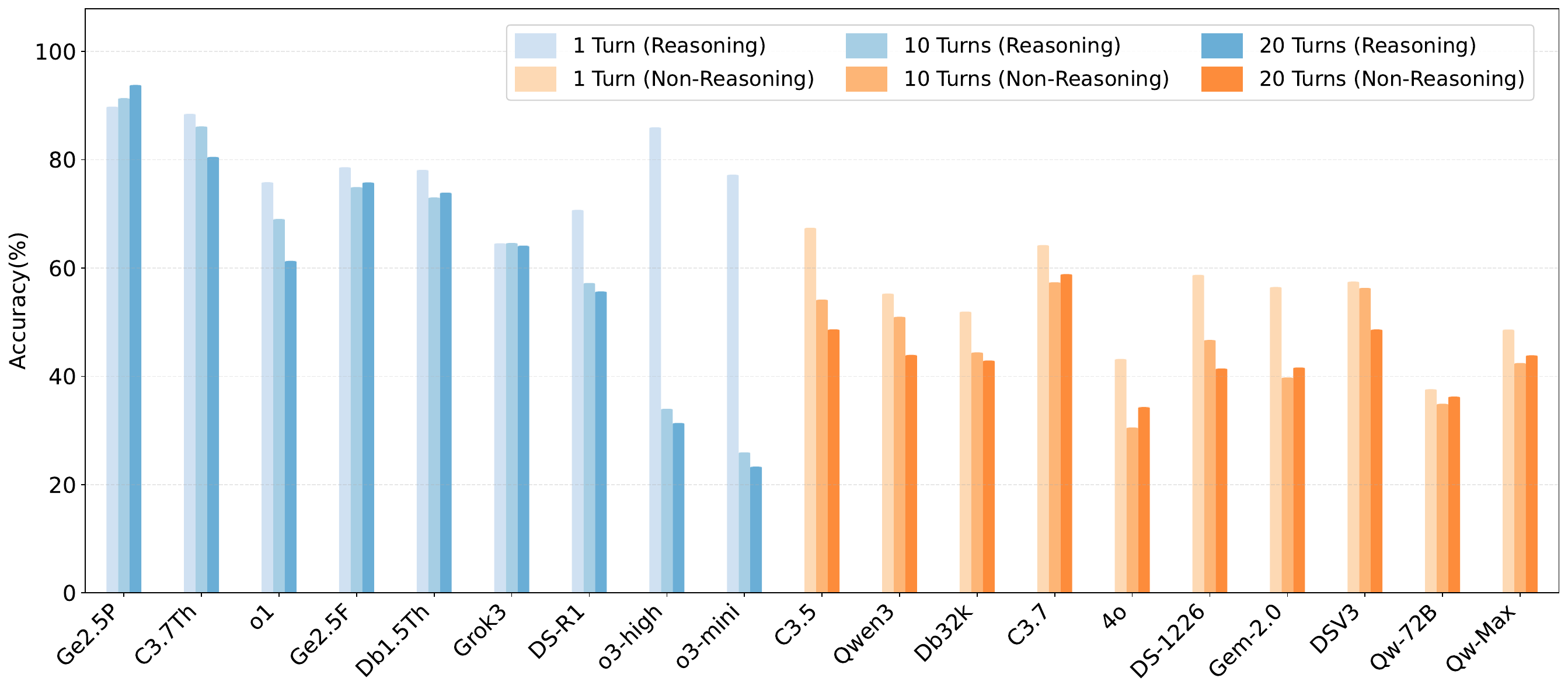}
    \caption{Detailed Results of Ablation on Turn Lengths with Identical Content}
    \label{fig:Ablation-cross turns}
\end{figure*}

\begin{figure*}[h!]
    \centering
    \includegraphics[width=\textwidth]{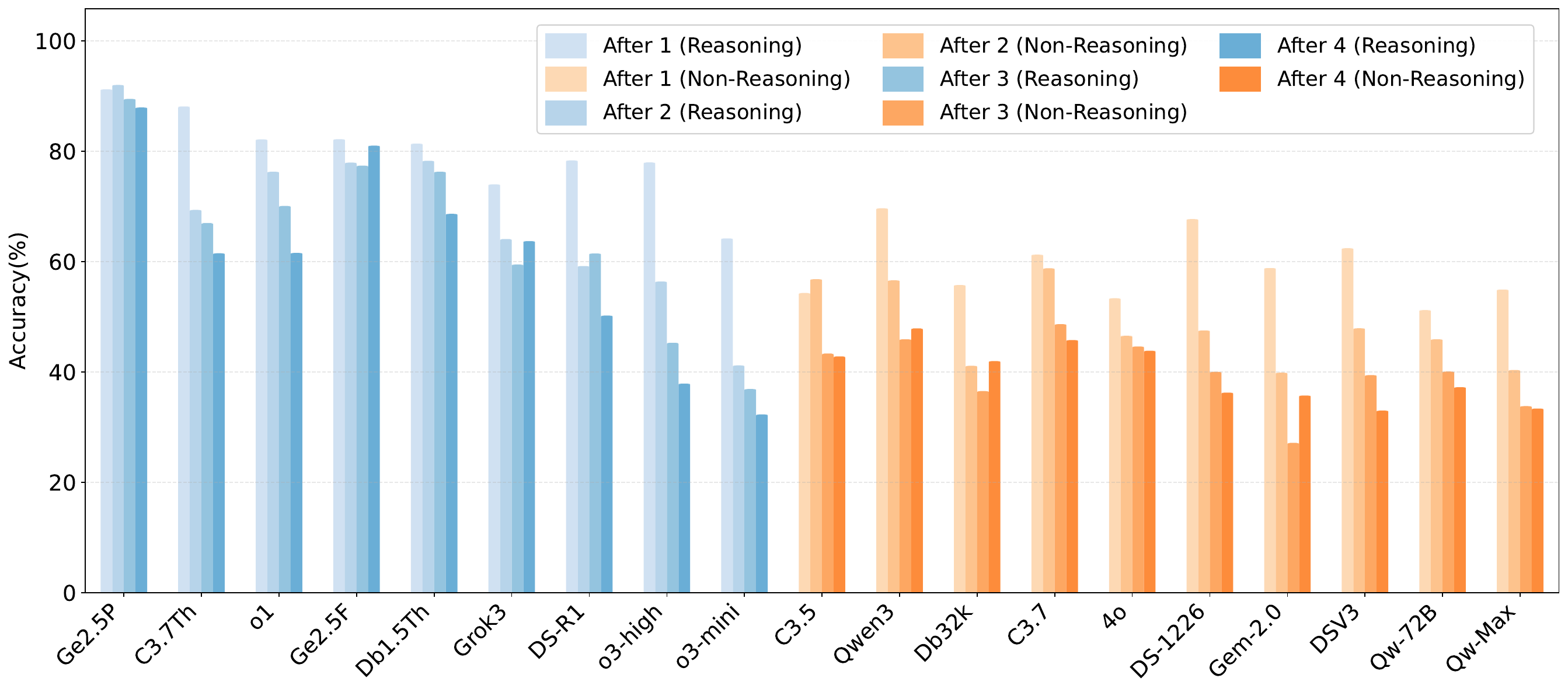}
    \caption{Detailed Ablation Results on Information Retrieval Distance}
    \label{fig:Ablation-cross distance}
\end{figure*}

In Section~\ref{subsec: rq2}, we analyzed the impact of cross-turn context on model performance from two dimensions: first, different turn lengths with the same content, and second, the retrieval distances between the queried information and the current turn. Representative experimental results of selected models were presented. In this section, we provide detailed visualization results for all models: the trend of model performance with respect to the turn of information retrieval is shown in Figure~\ref{fig:Ablation-cross turns}, while the trend of model performance relative to the information distance is illustrated in Figure~\ref{fig:Ablation-cross distance}.

\subsection{Ablation on Interaction Turns}
\begin{figure*}[h!]
    \centering
    \includegraphics[width=\textwidth]{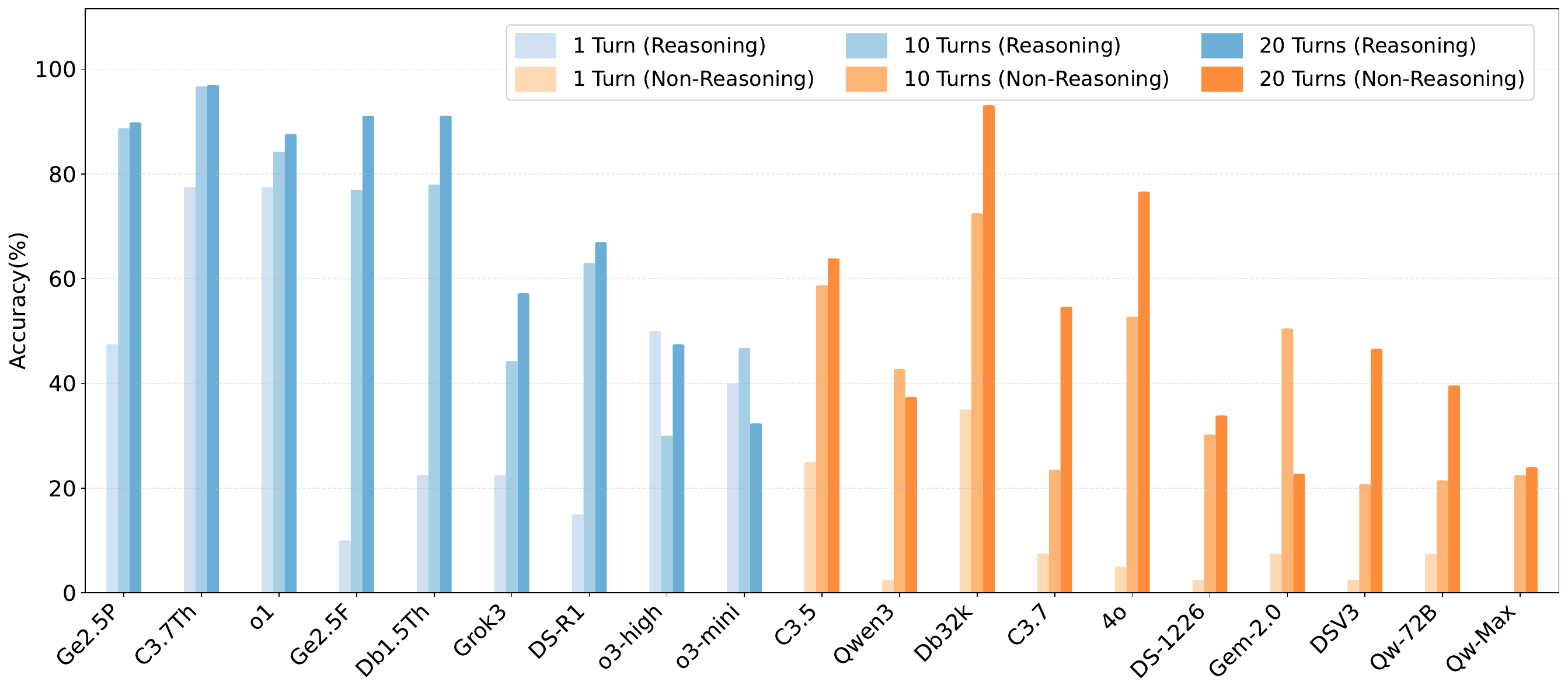}
    \caption{Detailed Ablation Results on Interaction Turns}
    \label{fig:Ablation-interaction}
\end{figure*}

In Section~\ref{subsec: rq3}, we analyzed the impact of interaction turns on model performance. Here, we provide detailed Figure~\ref{fig:Ablation-interaction} illustrating the results.

\newpage
\section{Task Categories}
\label{appendix: task categories}
This section provides the system prompts corresponding to four distinct task types and illustrative examples of questions, answers, and checklists for each associated subtask.

\subsection{System Prompt}
\label{appendix: system prompt}
The following is a basic example of a system prompt that can be directly applied to the Context Retrieval (CR), Information Reasoning (IR), and Task Switching (TS) tasks. However, the Instruction Following (IF) task introduces additional requirements built on this base prompt.
\begin{Prompt}[label={prompt:basic},title={System Prompt for CR, IR and TS Tasks}]

You have made plans with friends to watch a sports match, but your home TV is broken. You can only follow the match via text live streaming on your phone. During the match, you will chat based on the text updates. As a knowledgeable friend, you must:\\
\newline
- Accurately track match progress based on the text live stream;\\
- Naturally engage with your friend's comments and provide accurate answers;\\
- Ensure all quoted match data is absolutely correct;\\
- Only infer from objective live text information, no irrelevant content;\\
- Ensure your responses relate to your friend's comments, whether you're explaining, analyzing, or casually chatting, and always focus on the current match.\\
\newline
Your main goal is to match your friend's conversation rhythm and create an engaging, reliable dialogue experience.

---

Match: Phoenix Suns vs San Antonio Spurs

Team info: Phoenix Suns: Kevin Durant, Royce O'Neale, Nick Richards, Tyus Jones, Devin Booker, Ryan Dunn, Mason Plumlee, Bol Bol, Bradley Beal, Grayson Allen
San Antonio Spurs: Harrison Barnes, Bismack Biyombo, Chris Paul, De'Aaron Fox, Devin Vassell, Julian Champagnie, Jeremy Sochan, Keldon Johnson, Stephon Castle\\
---
\end{Prompt}

\begin{Prompt}[label={prompt: IF}, title = {System Prompt in Instrction Following}]

You have made plans with friends to watch a sports match, but your home TV is broken. You can only follow the match via text live streaming on your phone. During the match, you will chat based on the text updates. As a knowledgeable friend, you must:\\

- Accurately track match progress based on the text live stream;
- Naturally engage with your friend's comments and provide accurate answers;\\
- Ensure all quoted match data is absolutely correct;\\
- Only infer from objective live text information, no irrelevant content;\\
- Ensure your responses relate to your friend's comments, whether you're explaining, analyzing, or casually chatting, and always focus on the current match.\\

\textbf{Each reply must}:

- Be in English and less than 100 words;
- Be divided into three paragraphs, each ending with the same rhyme sound, and only three paragraphs are included, excluding the following tags.\\
- Begin with two parts on the same line:\\
- First, identify your friend's intent and print `[Update Score]` or `[Other Questions]`;\\
- Second, insert a tag according to this cycle:\\
    - Insert `[Tags A]` on the 1st, 5th, 9th... replies;\\
    - Insert `[Tags B]` on the 3rd, 7th, 11th... replies;\\
    - No tag on the 2nd, 4th, 6th, 8th... replies.\\

Your main goal is to match your friend's conversation rhythm and create an engaging, reliable dialogue experience.

---
Match: Phoenix Suns vs San Antonio Spurs\\

Team info: Phoenix Suns: Kevin Durant, Royce O'Neale, Nick Richards, Tyus Jones, Devin Booker, Ryan Dunn, Mason Plumlee, Bol Bol, Bradley Beal, Grayson Allen
San Antonio Spurs: Harrison Barnes, Bismack Biyombo, Chris Paul, De'Aaron Fox, Devin Vassell, Julian Champagnie, Jeremy Sochan, Keldon Johnson, Stephon Castle\\
---\\
\end{Prompt}

\subsection{Score Tracking Task}
\label{appendix: score tracking}
To simulate a realistic multi-turn dialogue, we define the core interaction between the user and the model as a text-based live broadcast excerpt from a game, provided by the user. The model is subsequently required to extract and report the score information of the teams. An illustrative example is provided below:

\begin{QuestionCase}[label={question: score tracking}]{Score Tracking Question}
\texttt{\large Question}: 
\\
\textbf{GAME INFO}

\textless{}Time-Play Table\textgreater{}

12:00\textbackslash{}tNick Richards vs. Bismack Biyombo (Tyus Jones gains possession)

11:47\textbackslash{}tKevin Durant misses 16-foot jumper

11:44\textbackslash{}tBismack Biyombo defensive rebound

11:30\textbackslash{}tDe'Aaron Fox makes 22-foot three point jumper (Chris Paul assists)

11:08\textbackslash{}tTyus Jones misses 22-foot three point jumper
\newline
\textbf{other play-by-play records......}
\newline
9:32\textbackslash{}tRoyce O'Neale defensive rebound

\textless{}/Time-Play Table\textgreater{}

---

Now, your friend has said:

What is the current score between the two teams? Can you help me analyze it?"\\
\newline
\texttt{\large Answer}:\\ "The current score is Phoenix Suns 2 - San Antonio Spurs 8."\\
\newline
\newline
\texttt{\large Checklist}:\\ \{

\quad    "Phoenix Suns's score is 2": 0.5,

\quad    "San Antonio Spurs's score is 8": 0.5

\}\\

\end{QuestionCase}

\subsection{Instruction Following}
\label{appendix: instruction following}
In the IF (Instruction Following) task, formatting requirements are derived from both the system prompt and the user’s in-dialogue instructions. At each turn, model responses are assessed for adherence to these requirements. Checklist scores are calculated based on the number of formatting constraints in the current question, with points distributed as evenly as possible across different sources. Each response is assigned a maximum score of 1.0.

\subsubsection{Fixed-format Single-turn Response}
In the Fixed-format Single-turn Response sub-task, the model is required to follow turn-specific instructions, which may include constraints on rhythm, formatting, or other stylistic aspects provided within the dialogue. 
\begin{QuestionCase}[label={question: IF single}]{Fixed-format Single-turn Response}
\texttt{\large Question}: \\
\textbf{GAME INFO}\\
\textless{}Time-Play Table\textgreater{}\\
3:40\textbackslash{}tJulian Champagnie defensive rebound\\
3:36\textbackslash{}tDevin Vassell misses running pullup jump shot\\
3:33\textbackslash{}tDevin Vassell offensive rebound\\
3:29\textbackslash{}tKeldon Johnson makes 23-foot three point jumper (Julian Champagnie assists)\\
3:14\textbackslash{}tGrayson Allen bad pass (Stephon Castle steals)\\
\newline
\textbf{other play-by-play records......}\\
\newline
1:51\textbackslash{}tSuns defensive team rebound\\
\textless{}/Time-Play Table\textgreater{}\\
---\\
Now, your friend has said:\\

I think Phoenix Suns is awesome, I want to write a sentence as support, can you help me write one? Write a sentence no longer than 20 words. Starting from the fifth word, the first letters of consecutive words must be sequentially composed of "Team name initial" plus "KING", and the order must match each letter, with no interruptions in between. If the team name is "NewYork Yankees", the first letters of consecutive words must be "NYKING".

\textbf{Two turns after this one}, after your normal answer, add a short summary in \textless{}Summary\textgreater{}\textless{}/Summary\textgreater{} tags. The summary should cover the match progress so far, under 50 words, placed at the end without affecting the answer's flow or format.\\
\newline
\texttt{\large Answer}: "Open answer, please evaluate according to the checklist."\\
\newline
\texttt{\large Checklist}: \{

\quad    "Adheres to English less than 100 words": 0.1,

\quad    "Adheres to three paragraphs, each ending with the same rhyme sound (excluding tags)": 0.1,

\quad    "Adheres to beginning with correct bracket ([Other Questions])": 0.05,

\quad    "Excludes tags as expected (No tag expected)": 0.25,

\quad    "Generated sentence starts the required letter sequence ('PSKING') from the 5th word": 0.25,

\quad    "Generated sentence uses consecutive words to form the complete required letter sequence ('PSKING')": 0.25

\}\\

\end{QuestionCase}

\subsubsection{Turn-conditioned Prompted Formatting}
Turn-conditioned Prompted Formatting is a sub-task where the response format is specified by the system prompt and remains consistent throughout the entire dialogue. To evaluate whether the model can distinguish between different dialogue turns, we assign different formatting requirements to different turns. The specific formatting instructions can be found in System Prompt~\ref{prompt: IF}. Moreover, in this sub-task, simpler formatting instructions (e.g., enclosing the response within a \textless{}Question\textgreater{} tag) are allocated lower scores, while more complex requirements (e.g., using distinct tags for different turns) are assigned higher weights to reflect their increased difficulty.

\begin{QuestionCase}[label={question: IF system}]{Turn-conditioned Prompted Formatting}
\texttt{\large Question}: \\
\textbf{GAME INFO}\\
\textless{}Time-Play Table\textgreater{}\\
12:00\textbackslash{}tNick Richards vs. Bismack Biyombo (Tyus Jones gains possession)\\
11:47\textbackslash{}tKevin Durant misses 16-foot jumper\\
11:44\textbackslash{}tBismack Biyombo defensive rebound\\
11:30\textbackslash{}tDe'Aaron Fox makes 22-foot three point jumper (Chris Paul assists)\\
11:08\textbackslash{}tTyus Jones misses 22-foot three point jumper\\
11:05\textbackslash{}tNick Richards offensive rebound\\
11:01\textbackslash{}tRoyce O'Neale misses 24-foot three point jumper\\
10:58\textbackslash{}tNick Richards offensive rebound\\
10:56\textbackslash{}tNick Richards misses dunk\\
10:56\textbackslash{}tDevin Vassell defensive rebound\\
10:48\textbackslash{}tChris Paul makes 15-foot pullup jump shot\\
10:37\textbackslash{}tKevin Durant misses 25-foot three point jumper\\
10:37\textbackslash{}tDe'Aaron Fox defensive rebound\\
10:24\textbackslash{}tHarrison Barnes makes 22-foot three point jumper (Chris Paul assists)\\
10:08\textbackslash{}tDevin Booker misses 13-foot pullup jump shot\\
10:08\textbackslash{}tSuns offensive team rebound\\
10:05\textbackslash{}tKevin Durant makes 14-foot pullup jump shot\\
9:48\textbackslash{}tDe'Aaron Fox misses driving floating jump shot\\
9:46\textbackslash{}tNick Richards defensive rebound\\
9:41\textbackslash{}tDevin Booker misses two point shot\\
9:39\textbackslash{}tChris Paul defensive rebound\\
9:36\textbackslash{}tHarrison Barnes misses 23-foot three point shot\\
9:32\textbackslash{}tRoyce O'Neale defensive rebound\\
\textless{}/Time-Play Table\textgreater{}\\
---\\
Now, your friend has said:\\
What is the current score between the two teams? Can you help me analyze it?\\
\newline
\texttt{\large Answer}: "Open answer, please evaluate according to the checklist"\\
\newline
\texttt{\large Checklist}: \{

  \quad  "Adheres to English less than 100 words": 0.2,
  
  \quad  "Adheres to three paragraphs, each ending with the same rhyme sound (excluding tags)": 0.2,
  
 \quad   "Adheres to beginning with correct bracket ([Other Questions])": 0.1,
 
 \quad   "Includes correct tag ([Tags A] expected)": 0.5
 
\}\\
\end{QuestionCase}

\subsubsection{Turn-conditioned Inferred Formatting}
Turn-conditioned Inferred Formatting focuses on evaluating whether the model complies with turn-specific formatting instructions that span multiple dialogue turns. Specifically, in one turn, the model may be instructed to append a summary of the game—wrapped in a \textless{}Summary\textgreater{} tag and limited to fewer than 50 words—two turns later, after the regular response. The checklist assigns individual scores to each of these requirements. Detailed instructions are provided within the multi-turn dialogues (see Question Case~\ref{question: IF single}), and examples of the questions and corresponding checklists are shown below:
\begin{QuestionCase}[label={question: IF turn}]{Turn-conditioned Inferred Formatting}
\texttt{\large Question}: \\
\textbf{GAME INFO}\\
\textless{}Time-Play Table\textgreater{}\\
End of Quarter / Half-time break\\
\textless{}/Time-Play Table\textgreater{}\\
---\\
Now, your friend has said:\\

Wow, the First quarter was really exciting! Both teams performed very well, and the game was very intense. I paid special attention to the top scorer(s) Stephon Castle, De'Aaron Fox, who seemed to have many impressive performances. Do you remember their/his shooting statistics for this quarter?

Please provide the answer in the following format for each top scorer:

[Player Name]:
Attempts: [Total Attempts]\\
Hits: [Total Hits]\\
- Two-pointers: [Number]\\
- Three-pointers: [Number]\\
- Free throws: [Number]\\
Total Points: [Total Points]\\

\texttt{\large Question}: "Open answer, please evaluate according to the checklist"\\
\newline
\texttt{\large Checklist}: \{

  \quad  "Adheres to English less than 100 words": 0.1,
  
   \quad "Adheres to three paragraphs, each ending with the same rhyme sound (excluding tags)": 0.1,
   
 \quad   "Adheres to beginning with correct bracket ([Other Questions])": 0.05,
 
 \quad   "Excludes tags as expected (No tag expected)": 0.25,
 
  \quad  "Includes \textless{}Summary\textgreater{} tags (Gate condition)": 0.0,
  
  \quad  "Summary content is less than 50 words": 0.25,
  
   \quad "Summary content accurately reflects game history up to this point": 0.25
   
\}\\
\end{QuestionCase}

\subsection{Context Retrieval}
\label{appendix: context retrieval}
The system prompt for CR task can be found in Prompt~\ref{prompt:basic}.

\subsubsection{Anchored Event Retrieval}
Anchored Event Retrieval requires the model to identify a specific event (e.g., a player's score) given a reference timestamp and a time interval. The evaluation assigns 0.5 points for correctly identifying the time and 0.5 points for retrieving the correct event. No points are awarded if the wrong player is identified.Each game period includes 2 such questions.

\begin{QuestionCase}[label={question: CR ancored},title = {Anchored Event Retrieval}]

\texttt{\large Question}: \\
    \textbf{GAME INFO}\\
    \textless{}Time-Play Table\textgreater{}\\
    End of Quarter / Half-time break\\
    \textless{}/Time-Play Table\textgreater{}\\
    ---\\
    Now, your friend has said:\\
    Wow, the Second quarter was really exciting! Both teams performed very well, and the game was very intense. \\
    I remember LeBron James scored in (quarter2 9:27), and 8.0 minutes and 38.9 seconds later, it seemed like someone had scored too. \\
    at what time did they/he score each time, and how many points did they/he score each time?\\

    Please list all the scoring information that meets the requirements

    **Note:** 
    - Set all NBA quarter breaks to 2 minutes.\\
    - If some players scored in the same time, please list all the scoring information.\\

    Please provide the answer in the following format for each top scorer:

    Player1 scored [Points] in [Time]\\
    Player2 scored [Points] in [Time]\\

    for example:\\
    Player1 scored 2 points in (quarter2 10:00)\\
    Player2 scored 3 points in (quarter3 12:00)\\
    ...\\

\texttt{\large Answer}: "Kris Dunn scored 3 in (quarter2 48.1)",\\

\texttt{\large Checklist}: \{

\quad  "Event 1: Kris Dunn scored 3": 0.5,
  
\quad  "Event 1: Kris Dunn scored in (quarter2 48.1)": 0.5,
  
\quad  "first\_check": \{
  
\quad \quad      "Incorrect player name(s) selected": 0
  
\quad  \}
\end{QuestionCase}

\subsubsection{Interval-based Event Retrieval}
Interval-based Event Retrieval requires the model to retrieve specific events occurring within a given time interval. The evaluation distributes the total score evenly across all target events within the specified interval. Each game period includes 1 such question.
\begin{QuestionCase}[label={question: CR interval},title = {Interval-based Event Retrieva}]

\texttt{\large Question}: \\
    \textbf{GAME INFO}\\
\textless{}Time-Play Table\textgreater{}\\
End of Quarter / Half-time break\\
\textless{}/Time-Play Table\textgreater{}\\
---\\
Now, your friend has said:\\

One of my friends was watching this NBA game too, but left to answer a phone call at (quarter1 4:29) and didn't return until (quarter1 2:45). 
Which goals did he miss? Please list the players who scored, when they scored, and how many points they got(including (quarter1 4:29) and (quarter1 2:45)).\\

Please provide the answer in the following format:

Player1 Name scored [Points] in [Time]
Player2 Name scored [Points] in [Time]\\
for example:\\
Player1 Name scored 2 points in (quarter2 10:00)\\
Player2 Name scored 3 points in (quarter3 12:00)\\
...\\

\texttt{\large Answer}: "Devin Vassell scored 2 in (quarter1 4:29)\\
Nick Richards scored 2 in (quarter1 4:12)\\
Nick Richards scored 1 in (quarter1 4:12)\\
Keldon Johnson scored 3 in (quarter1 3:29)\\
Stephon Castle scored 1 in (quarter1 3:12)\\
Stephon Castle scored 1 in (quarter1 3:12)\\
Stephon Castle scored 2 in (quarter1 2:45)"\\

\texttt{\large Checklist}: \{

  \quad  "Event 1: Devin Vassell scored 2 in (quarter1 4:29)": 0.14285714285714285,
  
  \quad  "Event 2: Nick Richards scored 2 in (quarter1 4:12)": 0.14285714285714285,
  
  \quad  "Event 3: Nick Richards scored 1 in (quarter1 4:12)": 0.14285714285714285,
  
 \quad   "Event 4: Keldon Johnson scored 3 in (quarter1 3:29)": 0.14285714285714285,
 
 \quad   "Event 5: Stephon Castle scored 1 in (quarter1 3:12)": 0.14285714285714285,
 
 \quad   "Event 6: Stephon Castle scored 1 in (quarter1 3:12)": 0.14285714285714285,
 
  \quad  "Event 7: Stephon Castle scored 2 in (quarter1 2:45)": 0.14285714285714285
  
\}\\

\end{QuestionCase}

\subsection{Information Reasoning}
\label{appendix: information reasoning}
The system prompt for IR task can be found in Prompt~\ref{prompt:basic}.

\subsubsection{Current Score Tracking}
Current Score Tracking shares the same question design as in the Multi-turn scenarios in the Question~\ref{question: score tracking}, but the question is accessed only at the end of the game.

\subsubsection{Score Lead Fluctuation Detection}
Score Lead Fluctuation Detection requires the model to identify the number of lead changes and the corresponding details during a specified game phase. The score is assigned as 0.2 for the correct count, and the remaining 0.8 is evenly distributed across the descriptions of each lead change. If there is at least one lead change in the period, one such question is included ; otherwise, none is included.

\begin{QuestionCase}[label={question: IR player},title = {Score Lead Fluctuation Detection}]

\texttt{\large Question}: \\
\textbf{GAME INFO}\\
\textless{}Time-Play Table\textgreater{}\\
End of Quarter / Half-time break\\
\textless{}/Time-Play Table\textgreater{}\\
---\\
Now, your friend has said:\\

In the First quarter, how many times did one team take the lead after previously being behind ? At what exact times did these lead changes occur, and which team became the new leader?  

**Note:** A tie does not count as changing the order. If a team that is lagging continuously goes from tying to leading, information on the time of the comeback needs to be recorded.

Please answer using the following format:

In the First quarter, the score order changed [Number] times. The specific time and the leading team are as follows:
At [Time1], [Team1]'s score had surpassed [Team2]'s score, ...\\

\texttt{\large Answer}: "In the First quarter, the score order changed 1 times. The specific time and the leading team are as follows:
11:30: San Antonio Spurs's score had surpassed Phoenix Suns's\\
"\\

\texttt{\large Checklist}: \{
    "The score order changed 1 times": 0.2,\\
    "At 11:30, San Antonio Spurs's score had surpassed Phoenix Suns's": 0.8\\
\}
\end{QuestionCase}

\subsubsection{Player Performance Impact Analysis}
Player Performance Impact Analysis requires the model to analyze the impact of a specified player’s on-court performance on a given segment of the game. For example, in the NBA game, the model is tasked with assessing how missed free throws or shots by the player affect the overall score during that segment. The scoring is evenly distributed across answers for each hypothetical scenario. If there is at least one lead change in the period, 2 such questions are included ; otherwise, one is included.
\begin{QuestionCase}[label={question: IR player},title = {Player Performance Impact
Analysis}]

\texttt{\large Question}: \\
\textbf{GAME INFO}\\
\textless{}Time-Play Table\textgreater{}\\
End of Quarter / Half-time break\\
\textless{}/Time-Play Table\textgreater{}\\
---\\
Now, your friend has said:\\

Wow, the Second quarter was really exciting! Both teams performed very well, and the game was very intense. 
Who has scored the most goals from (1 quarter 11:29) to (2 quarter 9:42)(including (1 quarter 11:29) and (2 quarter 9:42))? If he has not made any free throws during this period, which team should be leading now and by how many points? What if all of his goals during this period were not scored?\\

**Note:** if multiple players have the same score, please select the player whose name comes first in alphabetical order

Please provide the answer in the following format for the scorer:

The top scorer is [Player Name].
If he has not made any free throws during this period, the leading team should be [Team Name] and the score difference should be [Score Difference].\\
If all of his goals during this period were not scored, the leading team should be [Team Name] and the score difference should be [Score Difference].\\

\texttt{\large Answer}: "The top scorer is LeBron James.\\
If he has not made any free throws during this period, the leading team should be Los Angeles Lakers and the score difference should be 1.\\
If all of his goals during this period were not scored, the leading team should be LA Clippers and the score difference should be 8."\\

\texttt{\large Checklist}: \{

\quad    "Without free throws scenario: leading team is Los Angeles Lakers": 0.25,

\quad    "Without all goals scenario: leading team is LA Clippers": 0.25,

 \quad   "Without free throws scenario: The score difference is 1": 0.25,
 
 \quad   "Without all goals scenario: The score difference is 8": 0.25,
 
 \quad   "first\_check": \{
    
\quad \quad  "Incorrect player name(s) selected": 0

\quad    \}\\
\}\\
\end{QuestionCase}

\subsection{Task Switch}
\label{appendix: task-switch}
The system prompt for TS task can be found in Prompt~\ref{prompt:basic}.

\subsubsection{In-context Reasoning Query}
The In-context Reasoning Query in TS task are same as IR's Question.See Aeppendix~\ref{appendix: information reasoning}.

\subsubsection{Out-of-context Math Query}
Out-of-context Math Query mixes in math questions from MathQA within the dialogue. An example format is shown below.Each game period includes 3 such questions.
\begin{QuestionCase}[label={question: TS math}, title = {Example Question of Task Switch}]

\texttt{\large Question}:\\
\textbf{GAME INFO}\\
\textless{}Time-Play Table\textgreater{}\\
The competition is ongoing, users are chatting.\\
\textless{}/Time-Play Table\textgreater{}\\
---\\
Now, your friend has said:\\

I have a computer science problem to ask you. Please choose the correct answer from the following options and reply with the letter of the option directly:

Another term for out-of-distribution detection is?

A. precision-recall detection
B. underfitting detection\\
C. bias-variance tradeoff detection\\
D. regularization detection\\
E. one-class detection\\
F. overfitting detection\\
G. cross-validation detection\\
H. background detection\\
I. train-test mismatch robustness\\
J. anomaly detection\\
K. outlier detection\\
L. Both anomaly detection and outlier detection are correct\\
\newline
\texttt{\large Answer}: \\
"L"\\
\newline
\texttt{\large Checklist}: \{

\quad    "Correct answer is L": 1.0\\
\}\\
\end{QuestionCase}

\section{LLM Output Example}
\label{appendix:answer example}
\subsection{Case 1: Score Tracking and Analysis}

\begin{Prompt}[label={prompt:metric},title={Score Tracking's Answers}]

 \textbf{Question:} \{play-by-play table...\} What is the current score between the two teams? Can you help me analyze it?\\

\textbf{ Gemini-2.5-Pro:}

[1st turn]: The Mavericks are currently leading 4-0 against the Hornets. Moses Brown scored both baskets... (correct)

... (correct answers)

[Last turn]: End of 2nd Quarter! Score: Mavericks 50 - Hornets 49... (correct)\\

\textbf{o3-mini-high:}

[1st turn]: Right now, it's 4–0 in favor of the Mavericks. Moses Brown scored two points...

... (correct answer)

[3rd turn]: The current score is Dallas Mavericks 20, Charlotte Hornets 10... (incorrect)

... (incorrect answers)

[Last turn]: The score at the end of the second quarter is Dallas Mavericks 33, Charlotte Hornets 30... (incorrect)

\end{Prompt}

Both models completed score updates and analysis tasks, but the weaker o3-mini-high made errors early. Since the model tends to reference its previous responses rather than the original context, these initial errors propagated throughout subsequent turns, leading to error accumulation.

\subsection{Case 2: Context Retrieval}

\begin{Prompt}[label={prompt:metric},title={Context Retrieval's Answers}]

 \textbf{Question:} My friend left at (quarter2 4:09) and returned at (quarter3 5:07). Which goals did he miss? Please list players, timing, and points.\\

\textbf{Gemini-2.5-Pro:} 

Your friend missed a lot of action! Here are all the scores:

Dante Exum scored 3 points in (quarter2 4:09)\\
Dante Exum scored 2 points in (quarter2 3:24)\\
... (listed all correct answers)\\

\textbf{o3-mini-high:}

[Mark Williams] scored 2 points in (quarter3 11:31)

[Mark Williams] scored 2 points in (quarter4 0:13)\\
Explanation: In a standard NBA game, each quarter lasts 12 minutes... (incorrect calculation process)\\
Conclusion: From (quarter2 4:09) until (quarter3 5:07), there were no goals scored that he missed.\\

\end{Prompt}

The two models showed different performance on the same context retrieval task. Gemini-2.5-Pro correctly located the required content and provided accurate answers, while o3-mini-high incorrectly calculated time intervals, leading to wrong results.

\section{Metrics}
\label{appendix:metrics}

This section primarily describes the judge model, including its system prompt as well as input examples.

\begin{Prompt}[label={prompt:metric},title={System Prompt for the Judge Model}]
 
 You are a meticulous Grader tasked with evaluating the factual accuracy and completeness of a large language model's `prediction` regarding a sports game. Your evaluation must be based *strictly* on the provided `checklist`. The `reference\_answer` is provided as an ideal response example for context but scoring relies solely on the checklist items.\\

\#\#\# Evaluation Process:
1.  **Analyze Checklist:** Carefully examine the `checklist`. It contains specific facts or pieces of information (`"\{fact\}"`) expected in the prediction, each associated with points (`points`). It might also contain a `first\_check` section with overriding conditions.\\
2.  **Compare Prediction to Checklist:** For each item `"\{fact\}": points` in the main checklist:\\
    *   Determine if the specific `fact` is accurately present in the `prediction`. Reasonable paraphrasing is acceptable if the core meaning and data are identical to the fact stated in the checklist item.\\
    *   If the fact is present and correct in the prediction, award the corresponding `points`.\\
3.  **Calculate Initial Score:** Sum the points awarded for all correctly matched checklist items.\\
4.  **Apply `first\_check` Overrides (If Applicable):** Examine the `first\_check` section of the checklist, if present. This section typically contains key-value pairs like `\{"Reason for Zero Score": 0\}`.\\
    *   Evaluate if any of the listed "Reasons for Zero Score" accurately describe a fundamental flaw present in the `prediction`. A common example is if the prediction significantly misunderstands the core subject of the query (e.g., providing stats for the wrong player or team entirely when a specific one was asked about).\\
    *   If **any** condition listed in `first\_check` is determined to be true based on your assessment of the `prediction`, the **final score must be 0**. This overrides any points accumulated from the main checklist items.\\
5.  **Determine Final Score:** The final score is the summed points from the main checklist (Step 3), potentially overridden to 0 if a `first\_check` condition was met (Step 4). The score will be a float between 0.0 and 1.0.\\

\#\#\# Output Format:
Provide your evaluation in the following format:\\

[Scoring analysis]:
- Briefly explain which checklist items were found correctly in the `prediction` and which were missing or incorrect, referencing the awarded points.\\
- If a `first\_check` condition was met and resulted in a score of 0, **clearly state which specific condition** (the key from the `first\_check` dictionary, e.g., "Player name incorrect") was triggered and briefly explain why it applies to the prediction.\\
- Keep the analysis concise (around 100-150 words).\\
- Conclude with: "In conclusion, the prediction should receive x points" (where x is the final float score).\\

[Score]: x points

[JSON]:
```json\\
\{\\
  "answer\_score": [[score]]\\
\}\\
```\\
---\\
**Example Scenario (Illustrative):**\\

*Checklist:*
```json\\
\{\\
  "checklist": [
  
    \quad"Player A scored 10 points": 0.4,
    
   \quad "Player A had 3 assists": 0.3,
   
   \quad "Player B had 5 rebounds": 0.3
   
 \quad ],
  
\quad  "first\_check": [
  
    \quad \quad "Player name incorrect": 0
    
\quad  ]\\
\}\\

\end{Prompt}

\begin{Prompt}[label={prompt:input example},title={An Input Example for the Judge Model}]

\textless{}Prediction\textgreater{}\\
 Here's the detailed breakdown of lead changes in the first quarter:\\

In the First quarter, the score order changed **2 times**. The specific time and the leading team are as follows:  

1. **At 11:30**, the Suns' score (2 points) had surpassed the Spurs' score (0 points) after Kevin Durant's 14-foot pullup jump shot.  
   *(This was Phoenix's first and only lead until late in the quarter)*  \\

2. **At 5:46**, the Spurs' score (9 points) had surpassed the Suns' score (7 points) after Jeremy Sochan's driving dunk (assisted by Chris Paul).  
   *(San Antonio then maintained the lead for the rest of the quarter, though Phoenix briefly tied it at 16-16 before falling behind again)*  \\

**Note:** The Suns briefly tied the game at 16-16 (4:12) but did not retake the lead, so this does not count as a lead change. The only two clear lead changes occurred at the times listed above.\\
\textless{}/Prediction\textgreater{}\\
\newline
\textless{}Reference Answer\textgreater{}\\
In the First quarter, the score order changed 1 times. The specific time and the leading team are as follows:\\
11:30: San Antonio Spurs's score had surpassed Phoenix Suns's\\
\textless{}/Reference Answer\textgreater{}\\
\newline
\textless{}Checklist\textgreater{}\\
\{"The score order changed 1 times": 0.2, "At 11:30, San Antonio Spurs's score had surpassed Phoenix Suns's": 0.8\}\\
\textless{}/Checklist\textgreater{}\\
\end{Prompt}

\begin{Prompt}[label={prompt:output example},title={An Output Example for the Judge Model}]

[Scoring analysis]:\\

The prediction incorrectly states that the score order changed 2 times, whereas the checklist requires it to be 1 time (0 points awarded for this item). However, the prediction correctly identifies that at 11:30, the San Antonio Spurs's score had surpassed the Phoenix Suns's (0.8 points awarded for this item). The explanation provided in the prediction aligns with the checklist's requirement for the specific time and leading team. No 'first\_check' conditions are present in the checklist, so no overrides apply.\\

In conclusion, the prediction should receive 0.8 points.\\

[Score]: 0.8 points\\

[JSON]:\\

```json\\

\{

  "answer\_score": [[0.8]]

\}

```
\end{Prompt}

\section{Impact of Model Scale on Multi-turn Dialogue Performance}
\label{appendix:model_size_analysis}
\begin{table}[!ht]
    \centering
        \caption{Performance of Qwen3 models of varying sizes on multi-turn dialogue tasks in \ourbenchmark.}
    \begin{tabular}{lccccc}
    \toprule
        \textbf{Model} & \textbf{Overall} & \textbf{IF} & \textbf{CR} & \textbf{IR} & \textbf{TS} \\
    \midrule
        Qwen3-32B & 30.96 & 38.35 & 27.82 & 30.92 & 26.74 \\
        Qwen3-14B & 28.27 & 42.15 & 21.26 & 24.84 & 24.83 \\
        Qwen3-8B & 27.12 & 45.69 & 17.38 & 22.36 & 23.05 \\
        Qwen3-4B & 25.52 & 46.77 & 16.62 & 17.62 & 21.07 \\
    \bottomrule
    \end{tabular}
    \label{tab:qwen3_model_scale}
\end{table}

The results indicate a clear correlation between model size and performance across all tasks. Larger models consistently achieve higher overall scores and demonstrate superior capabilities in instruction following (IF), context retrieval (CR), information reasoning (IR), and task switching (TS), whereas smaller models exhibit marked performance degradation, particularly in tasks that require cross-turn reasoning and interactive adaptation. These findings highlight the critical role of model capacity in effectively handling complex multi-turn dialogue scenarios.

\section{Results on CoT Prompting}
\label{appdendix:CoT}
\subsection{CoT Prompt}
\label{appdendix:CoT Prompt}
\begin{Prompt}[label={prompt:basic},title={CoT Prompt}]

You should think the question step by step and then response the question, there are some templates you should follow:\\
<Thinking>here is your thinking process</Thinking>\\
<Response>here is your response</Response>\\
\\
here is an example:\\
Question: What is the current score between the two teams?\\
<Thinking>\\
1. First, there were 5 goals in total: Goal1 to Goal5\\
2. Second, the scorers were A, B, C, A, D, each worth 1 point\\
3. Then, their teams are X, Y, X, X, Y respectively\\
4. Finally, calculate scores: TeamX 3pts, TeamY 2pts\\
</Thinking>\\
<Response>The current score is TeamX 3 : TeamY 2.</Response>\\
\end{Prompt}
\subsection{Result on CoT Prompting}

\begin{table}[h]
\centering
\caption{Effect of CoT prompting on the overall performance of non-reasoning models.}
\label{tab:model_comparison}
\begin{tabular}{lccc}
\toprule
\textbf{Model} & \textbf{CoT} & \textbf{Zero-shot} & \textbf{Diff} \\
\midrule
Gemini-2.0-Flash & 40.74 & 35.61 & 5.13 \\
Claude-3.5-Sonnet & 55.74 & 43.17 & 12.57 \\
GPT4o-1120 & 40.03 & 35.83 & 4.20 \\
Llama4-Maverick & 35.05 & 34.50 & 0.55 \\
DeepSeek-V3-1226 & 43.43 & 33.16 & 10.27 \\
DeepSeek-V3-0324 & 47.23 & 37.31 & 9.92 \\
GPT-4.1-mini-0414 & 33.08 & 31.23 & 1.85 \\
Qwen3-32B & 29.54 & 30.96 & -1.42 \\
Qwen2.5-Max & 34.19 & 30.41 & 3.78 \\
Qwen3-14B & 26.16 & 28.27 & -2.11 \\
Llama4-Scout & 21.49 & 27.27 & -5.78 \\
\bottomrule
\end{tabular}
\end{table}

The application of chain-of-thought prompting generally improves the performance of non-reasoning models, though the magnitude of these gains remains closely dependent on the underlying model capacity. As shown in Table~\ref{tab:model_comparison}, \texttt{Claude-3.5-Sonnet} achieves the highest overall score of 55.74, with an improvement of +12.57 points over its zero-shot baseline, while both \texttt{DeepSeek-V3} variants exhibit increases exceeding 9.92 points. In contrast, the effect is not consistent across models: \texttt{Qwen3-14B} and \texttt{Llama4-Scout} even show declines in overall performance, reflecting limited capacity for structured multi-step reasoning. These results demonstrate that although CoT prompting can enhance systematic processing in some non-reasoning models, fundamental architectural capacities remain the primary determinant of overall performance.


\section{Attention Visualization Analysis}
\label{appendix:attention}

To examine how input structure influences the distribution of model attention, we perform a visualization-based analysis of \texttt{Qwen2.5-7B-Instruct} during generation. We extract multi-head attention weights from the top four transformer layers, average across heads within each layer, and subsequently aggregate across layers to obtain a unified attention map for each generation step. An attention score for each input token is then defined as the average amount of attention it receives throughout the generation process. For any contiguous input span, its total attention score is computed as the sum over its constituent tokens.

Formally, let the input sequence be $X = \{x_1, x_2, \dots, x_n\}$ and the generated sequence be $Y = \{y_1, y_2, \dots, y_T\}$. At each generation step $t$, the model produces an attention distribution $\mathbf{A}^{(t)} \in \mathbb{R}^n$, indicating the attention from $y_t$ to each $x_i$. The attention score for token $x_i$ is computed as:

\[
\text{Score}(x_i) = \frac{1}{T} \sum_{t=1}^{T} \mathbf{A}^{(t)}_i
\]

For a text span $S \subseteq X$, the total attention score is:

\[
\text{Score}(S) = \sum_{x_i \in S} \text{Score}(x_i)
\]

We apply this method to two contrastive experimental conditions to investigate the impact of the input structure.

\paragraph{Effect of Turn Count on Attention Focus.}
We segment the same play-by-play input into 1-turn, 10-turn, and 20-turn dialogue formats and evaluate the attention received by semantically relevant spans. We observe that as the number of turns increases, attention to key content consistently declines. This pattern aligns with the performance degradation reported in the main text, suggesting that increased turn-based fragmentation may impair the model’s ability to effectively integrate context.

\paragraph{Attention Overhead from Special Tokens.}
We further compare a single-turn concatenation with a 20-turn dialogue input containing identical content. In the multi-turn setting, the number of special tokens (e.g., \texttt{<|im\_end|>}) increases substantially, from 4 to 85. These tokens appear to receive disproportionately high attention, which may dilute the model’s focus on semantically meaningful input, potentially contributing to reduced task performance.

Visualization is rendered as highlighted text, where color intensity reflects linearly normalized attention scores. We present only the top 1500 input tokens ranked by score and exclude whitespace and control characters to preserve only visible content. All generations are performed using fixed sampling parameters: temperature = 0.7, top-p = 0.9, and top-k = 40.

Figure~\ref{fig:aggregation_attention} shows attention disproportionately allocated to special tokens, while Figure~\ref{fig:turn_attention_qwen} demonstrates a drop in attention to key content from 0.000714 (1 turn) to 0.000274 (20 turns), representing a 2.6× reduction. These results further support the interpretation that input structure affects attention distribution and may influence overall model behavior.

\end{document}